\documentclass{article}

\usepackage{arxiv}

\usepackage[utf8]{inputenc} 
\usepackage[T1]{fontenc}    
\usepackage{hyperref}       
\usepackage{url}            
\usepackage{booktabs}       
\usepackage{amsfonts}       
\usepackage{nicefrac}       
\usepackage{microtype}      
\usepackage{lipsum}
\usepackage{times}
\usepackage{soul}
\usepackage{url}
\usepackage{graphicx}
\usepackage{amsmath}
\usepackage{booktabs}
\usepackage{soul}
\usepackage{array}
\usepackage{microtype}
\usepackage{subfigure}
\usepackage{amsfonts}
\usepackage{amssymb}
\usepackage{xcolor}
\usepackage{enumitem}
\usepackage{algorithm}
\usepackage{algorithmic}
\usepackage{amsthm}
\usepackage[symbol]{footmisc}
\usepackage[normalem]{ulem}
\newtheorem{theorem}{Theorem}

\title{Adversarial Deep Embedded Clustering:\\on a better trade-off between\\Feature Randomness and Feature Drift}

\author{
  Nairouz Mrabah\\
  Department of Computer Science\\
  University of Quebec at Montreal\\
  Montreal, QC, Canada\\
  \texttt{mrabah.nairouz@courrier.uqam.ca} \\
   \And
  Mohamed Bouguessa\\
  Department of Computer Science\\
  University of Quebec at Montreal\\
  Montreal, QC, Canada\\
  \texttt{bouguessa.mohamed@uqam.ca} \\
   \And
  Riadh Ksantini\\
  Department of Computer Science\\
  University of Windsor\\
  Windsor, ON, Canada\\
  \texttt{ksantini@uwindsor.ca} \\
}

\begin{document}
\maketitle

\begin{abstract}
Clustering using deep autoencoders has been thoroughly investigated in recent years. Current approaches rely on simultaneously learning embedded features and clustering the data points in the latent space. Although numerous deep clustering approaches outperform the shallow models in achieving favorable results on several high-semantic datasets, a critical weakness of such models has been overlooked. In the absence of concrete supervisory signals, the embedded clustering objective function may distort the latent space by learning from unreliable pseudo-labels. Thus, the network can learn non-representative features, which in turn undermines the discriminative ability, yielding worse pseudo-labels. In order to alleviate the effect of random discriminative features, modern autoencoder-based clustering papers propose to use the reconstruction loss for pretraining and as a regularizer during the clustering phase. Nevertheless, a clustering-reconstruction trade-off can cause the \textit{Feature Drift} phenomena. In this paper, we propose ADEC (Adversarial Deep Embedded Clustering) a novel autoencoder-based clustering model, which addresses a dual problem, namely, \textit{Feature Randomness} and \textit{Feature Drift}, using adversarial training. We empirically demonstrate the suitability of our model on handling these problems using benchmark real datasets. Experimental results validate that our model outperforms state-of-the-art autoencoder-based clustering methods.
\end{abstract}

\keywords{Unsupervised Learning, Deep Learning, Clustering, Autoencoders.}

\section{Introduction} \label{S:1}

The main focus of clustering is to partition the original data into clusters without using any supervisory signal. During the last decades, a plethora of clustering algorithms has been proposed to overcome three main challenges. The first challenge is the \textit{high-dimentionality} of the exiting real-world information. For example, a typical image has thousands of pixels. This characteristic makes the clustering task more difficult due to the well-known curse of dimensionality \cite{paper56}. The \textit{large amount} of data or big data, as popularized by the public community \cite{paper57}, constitutes the second challenge. Computationally, processing large-scale datasets is generally associated with time and memory overheads. Last but not least, the \textit{high-semantic} aspect of natural data makes clustering a more challenging task. For example, two images of cats may look nothing like each other from pixel-level although they belong to the same class. The high-semantic aspect of natural data can be explained by the compositional hierarchy of features \cite{paper20}. In that respect, it is well-know that high-level features are nothing than a combination of lower-level ones. To give an example in the case of visual data, edges are combined to construct motifs, which are the keystone for building objects. The same applies to speech and text datasets.

Classical clustering methods, such as, k-means \cite{paper14}, Gaussian Mixture Model \cite{paper47}, DBSCAN \cite{paper48} and Mean Shift \cite{paper15} are shallow models. They rely on computing distance-based similarities in the raw data-space or in the space where the handcrafted features live. However, features engineering is task-specific. Therefore, it is inappropriate to integrate such pre-processing task in the pipeline of a general-purpose clustering framework. Moreover, natural data (e.g., images and videos) have high-dimensional and high-semantic aspects. So, when dealing with such datasets, the conventional clustering approaches have limited performance as the computational time increases considerably. In addition, distance-based metrics computed in the raw data-space are unreliable for discovering semantic similarities. 


To address the curse of dimensionality, the original high-dimensional data should be projected in a low-dimensional feature space. While abundant literature revolves around unsupervised dimensionality reduction, there are two main families. The first one consists of the linear dimensionality reduction methods, such as, Principal Component Analysis \cite{paper49} and Factor Analysis \cite{paper50}. The second family is based on the assumption that the most pertinent information lies on a low dimensional manifold (not a linear subspace) \cite{paper52}. Multi-dimensional scaling \cite{paper51}, Isometric Feature Mapping \cite{paper52} and Hessian Eigen mapping \cite{paper53} are among the popular manifold dimensionality reduction techniques. Although the linear and non-linear methods aim to preserve substantial information, they are prone to discriminative information loss, which in turn decimates the clustering performance. 

Projected clustering \cite{paper82} and subspace clustering \cite{paper83} have gained popularity thanks to their ability to address the problem of high-dimensional data clustering, as they identify relevant dimensions that exhibit the cluster structure. Unlike pure dimensionality reduction techniques, these methods do not ignore the discriminative aspect. Yet, they are only effective when the data meets the linear subspace hypothesis, which is rarely the case for natural data. 

Differently, kernel k-means \cite{paper54} and spectral clustering \cite{paper12} map the data to non-linear manifolds. Nevertheless, the transformation capacity of these approaches is limited. They generally underfit the complexity and semanticity of real-world information. Added to that, their computational time usually grows considerably when processing large databases. 

The recent advancement in unsupervised representation learning \cite{paper17,paper5} based on deep neural networks gave birth to a new family of clustering strategies, known as deep clustering. The multi-layers architecture has become the natural choice when it comes to processing large, high-semantic and high-dimensional datasets for several reasons. First, backpropagation and Stochastic Gradient Descent (SGD) allow to update the network weights in a cheap way without the need to loop around the whole dataset, in every single iteration. That's why a neural network is well-adopted for analyzing large data. Second, the compositional nature of the data \cite{paper20} justifies the need to gradually extract higher semantic representations from one layer to another using non-linear projection. Finally, the number of neurons in the hidden layers defines the dimensionality of the embedded spaces. Hence, a deep architecture allows to reduce dimensionality if it is designed to have a low-dimensional latent space. In spite of the deep learning success in many supervised applications, leveraging the power of neural networks in performing data clustering is still an open problem.  


The most prominent deep clustering approaches rely on autoencoders \cite{paper27,paper28,paper29,paper30,paper31}. Some other deep clustering strategies harness an encoding architecture \cite{paper21,paper22,paper23,paper24,paper25,paper26} without a decoding network. However, dispensing with the decoder and clustering the data in a latent space just using pseudo-labels can mislead the training process. This is because pseudo-labels are primarily generated based on \textit{hypothetical} similarities, which generally underfit the semanticity of natural datasets. We call this problem \textit{Feature Randomness}. The rational of choosing auto-encoding as the standard deep clustering architecture can be imputed to the limited reliability of pseudo-supervision when used alone. The reconstruction allows to rebuild the input samples after encoding them in a low dimensional latent space. The ability to reconstruct a point from a low-dimensional representation suggests that the latent space captures the key factors of variations and similarities. Otherwise, it would be impossible to regenerate the data samples.

Autoencoder-based clustering approaches generally consists of a joint optimization process. The reconstruction cost is combined with a clustering-specific objective function.
Referring to the previous point, retaining the reconstruction cost during the clustering phase helps in reducing \textit{Feature Randomness} by blocking the encoder from generating random discriminative features. However, regularizing with the reconstruction end-to-end can lead to \textit{Feature Drift}. This problem emerges due to the natural trade-off between clustering and reconstruction. Put it differently, while latent clustering allows to group and separate the embedded samples by emphasizing within-cluster similarities and destroying between-cluster similarities, the reconstruction is associated with preserving all factors of similarities.

Meanwhile, Generative Adversarial Network (GAN) \cite{paper1} has shown great promise in learning complex natural data distributions. It allows to synthesize out-of-sample data points. Besides, it has been shown that GAN can obtain images with impressive visual quality \cite{paper18,paper32}. Apart from being a successful generative model, the adversarial training strategy has inspired several modern achievements on unsupervised representation learning \cite{paper5,paper6,paper7,paper9}. Although GAN does not come with an encoder out-of-the-box, some recent papers have suggested to extend the classical GAN framework to permit data encoding in a latent space, where the semantic factors of variations and similarities are better-emphasized \cite{paper6,paper7,paper10,paper8}. Nevertheless, it is still unclear to what extent the features learned, based on a deep generative model, can be useful for downstream discriminative tasks (e.g., classification and clustering). 

To address the aforementioned problems, we propose Adversarial Deep Embedded Clustering (ADEC). Our framework consists of eliminating the strong competition between embedded clustering and reconstruction without incurring a \textit{Feature Randomness} cost. This is done by getting the strong competition outside of a single network, while relying on a discriminator, in order to make sure that the embedded representations preserve the intrinsic data characteristics. Optimizing every objective function in a separate network , based on adversarial training, allows to reach a better trade-off between \textit{Feature Randomness} and \textit{Feature Drift}. Experimentation on real benchmark datasets shows the superiority of our framework, in terms of accuracy (ACC) and normalized mutual information (NMI). In a nutshell, the key contributions of this paper are:

\begin{itemize}
  \item Devising a new pretraing framework based on adversarially constrained interpolation and data transformation.
  \item Overcoming the clustering-reconstruction compromise based on an adversarial training strategy.
  \item Enhancing state-of-the-art autoencoder-based clustering by alleviating \textit{Feature Randomness} and \textit{Feature Drift}.
   \item Outperforming modern clustering models in terms of ACC and NMI on real benchmark datasets.
\end{itemize}

The rest of this paper has the following organization: Section 2 is devoted to related work. Section 3 presents an analysis of the trade-off between \textit{Feature Randomness} and \textit{Feature Drift}. In section 4, we present our methodology for tackling the \textit{Feature Randomness} and \textit{Feature Drift} problems. In Section 5, we show our experimental results. Finally, section 6 concludes the paper. 

\section{Related work}
To demonstrate the merit of our proposed framework, we provide a critical review of mainstream approaches related to our work. ADEC comes under the realm of deep clustering strategies. Furthermore, it is deemed to be part of the concerted effort to combine GANs and autoencoders. To this end, we shall review the existing deep clustering approaches and the unifying techniques of GANs and autoencoders. We should also review DEC \cite{paper27} and IDEC \cite{paper28} since they constitute our principal baselines. 


\begin{figure*}[ht]
\vskip 0.2in
\begin{center}
\centerline{\includegraphics[width=400pt, height=60pt]{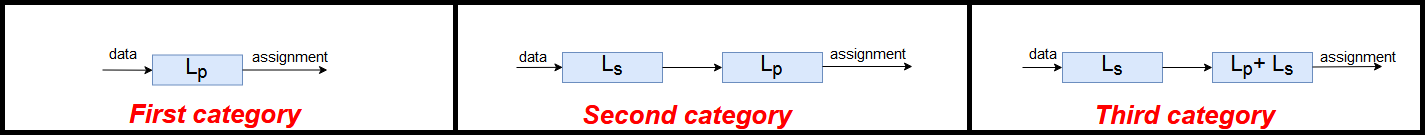}}
\caption{The different deep clustering categories. The self-supervised loss $L_{s}$ enforces reasonable general-purpose features and the pseudo-supervised loss $L_{p}$ is used for clustering the embedded data.}
\label{fig:deep_clustering_categories}
\end{center}
\end{figure*}

\subsection{Deep Clustering}
In deep unsupervised learning, typically, there are two conceivable options to make up for the absence of supervisory signals. The first option consists of contriving a pretext task that encourages to learn general-purpose features. It is better known as self-supervision. For this case, labels are extracted from the input data. The intuition behind self-supervision is that the pretext task can not be solved efficiently without gaining a semantically high-level grasp of the input data. The obtained features can be used to outsource downstream tasks, such as, classification. There exists a wide variety of pretext tasks. Among them, the vanilla reconstruction, the denoising loss \cite{paper58}, the variational loss \cite{paper64}, the adversarial loss \cite{paper1}, predicting the location of image patches \cite{paper59}, predicting the permutations of a "jigsaw puzzle"\cite{paper61}, predicting unpainted image patches\cite{paper62}, and predicting image colorization \cite{paper63}. The second option consists of contriving a pseudo supervisory signal. Similar to self-supervision, the labeling is available within the data. However, for pseudo-supervision, labels are predicted. Therefore, they are not 100\% correct. In this paper, $L_{s}$ denotes a self-supervised loss and $L_{p}$ denotes a pseudo-supervised loss. 

A possible categorization of deep clustering methods can be imputed to the used loss functions and the way they are combined. Based on that, the existing models fall into three main categories. In Figure \ref{fig:deep_clustering_categories}, the framework of each category is illustrated. Within the actual context, the pseudo-supervised loss stands for the embedded clustering objective function, which can be any one of the typical clustering objective functions, such as, Gaussian Mixture Model (GMM) or k-means. As regards the self-supervised cost, reconstruction is commonly selected.

For the first category, the clustering is directly performed using a pseudo-supervised loss. However, the self-acquired labels are unreliable due to their hypothetical aspect. This can mislead the data grouping by learning non-representative features, which in turn deteriorates the discriminative ability of the model. Concisely, the main weakness of the methods affiliated with this category is \textit{Feature Randomness}. As part of this category, \textit{Yang et al.} proposed JULE \cite{paper22}, a deep recurrent framework that allows to perform agglomerative clustering and feature learning alongside with a unified triplet loss. The whole process is optimized end-to-end. However, one among the prominent downsides of JULE is the run-time overhead due to the recurrent framework. \textit{Chang et al.} proposed DAC \cite{paper23}, a framework that enables to cluster the data based on pairwise constraints. DAC has curriculum learning strategy, where only high-confidence training samples are selected. In another line of research, \textit{Hu et al.} proposed IMSAT \cite{paper26}, which consists of maximizing the mutual information between discrete predictions and their associated data points. The loss function of IMSAT is regularized by a self-augmented training term that allows to penalize the discrepancy between initial data and their geometrically transformed ones. DeepCluster \cite{paper24} is another framework intimately tied to this category. It alternates between two basic steps. First, the latent representations are clustered by k-means. Then, the obtained clustering assignments are fed to a convolutional neural network as supervisory signals to learn better features. It was mainly applied to large-scale datasets.

As for the second category, the network is pretrained using a self-supervised cost function. Then, the obtained latent features are finetuned by retraining using pseudo labels. Compared with the random initialization, pretraining with a self-supervised loss leads to improved initial features. Nevertheless, there is no correction mechanism to attenuate the noisy labels harm. Hence, \textit{Feature Randomness} is a strongly remaining concern for this category. As part of this category, DEC \cite{paper27} is the first deep clustering framework to follow a pretraining-finetuning strategy.

The third category consists of pretraining using a self-supervised cost function similar to the second category. However, their finetuning phases are different. In fact, the third category regularizes the pseudo-supervised objective function with a self-supervised one. The advantage of such a strategy is that it offers a mechanism to reduce \textit{Feature Randomness}. However, combining pseudo-supervision and self-supervision can lead to a strong competition between them. In order to balance the two cost functions, a hyperparameter is required. To give an example, \textit{Guo et al.} proposed IDEC \cite{paper28} and  \textit{Dizaji et al.} proposed DEPICT \cite{paper36}. Both models can be considered as extensions to DEC. They regularize the clustering loss with reconstruction during the finetuning phase. Therefore, the decoder is maintained throughout the whole training process. The main difference between them is that DEPICT utilizes a convolutional architecture, whereas IDEC leverages a fully-connected autoencoder. Apart from that, \textit{Yang et al.} proposed DCN \cite{paper29}. Compared with IDEC and DEPICT, DCN optimizes a latent k-means objective function. VaDE \cite{paper35} is another model from this category. Its variational auto-encoding architecture allows to impose a GMM latent distribution. Thanks to the reparameterization trick, VaDE can be optimized using backpropagation. In addition to clustering, VaDE can generate data samples. Even so, it is subjected to elevated computational complexity similar to all the other variational frameworks.



\subsection{Deep Embedding Clustering (DEC)}
DEC \cite{paper27} has two phases. The pretraining phase allows to learn low-dimensional embedded representations by training the autoencoder with vanilla reconstruction. Then, comes the clustering phase. First, the decoder is discarded. After that, the encoder is trained to jointly optimize the embedded representations and the clustering centers. For every training iteration, a soft clustering assignment $q_{ij}$ (\ref{eq:q_ij}) is computed based on the Student’s t-distribution. $q_{ij}$ represents an assessment of the  between the embedded data point $z_{i}$ and the center $\mu_{j}$.

\begin{equation} \label{eq:q_ij}
  \begin{aligned}
    q_{ij} = \frac{(1 + \frac{\left \| z_{i} - \mu_{j} \right \|^{2}}{\alpha })^{-\frac{\alpha+1}{2}}}{\sum_{j'}(1 + \frac{\left \| z_{i} - \mu_{j'} \right \|^{2}}{\alpha })^{-\frac{\alpha+1}{2}}}.
  \end{aligned}
\end{equation}

The DEC loss function (\ref{eq:L_DEC}) is the Kullback Leibler divergence between the soft clustering assignment $q_{ij}$ and an auxiliary target distribution $p_{ij}$ (\ref{eq:p_ij}).

\begin{equation} \label{eq:L_DEC}
  \begin{aligned}
    L_{DEC} = KL(P||Q) = \sum_{i} \sum_{j}p_{ij} log(\frac{p_{ij}}{q_{ij}}),  
  \end{aligned}
\end{equation}

\begin{equation} \label{eq:p_ij}
  \begin{aligned}
    p_{ij} = \frac{q_{ij}^{2} / f_{j}}{\sum_{j'} q_{ij'}^{2} / f_{j'}}.
  \end{aligned}
\end{equation}

\subsection{Improved Deep Embedded Clustering (IDEC)}
IDEC \cite{paper28} has the same pretraining phase as DEC. The main difference between them is that IDEC is finetuned to minimize joint embedded clustering and reconstruction as described by (\ref{eq:L_IDEC}).

\begin{equation} \label{eq:L_IDEC}
  \begin{aligned}
    L_{IDEC} = L_{r} + \gamma L_{DEC}.
  \end{aligned}
\end{equation}

$L_{r}$ stands for the reconstruction and $(\gamma >0)$ is in charge of balancing the two costs. The key idea of IDEC is to block the clustering loss from corrupting the feature space. However, we argue that combining embedded clustering and vanilla reconstruction gives birth to a strong competition between them (i.e., \textit{Feature Drift}). 

\subsection{Combining Autoencoders with GANs}
Interpolating data samples from the prior distribution, in the latent space of the generator, leads to realistic and semantically explainable variations \cite{paper18,paper32}. As a consequence of GAN effectiveness in capturing the semantic factors of variations, many researchers have studied the inverse mapping problem (i.e., projecting the data back in the embedded space) \cite{paper7,paper10,paper6,paper8}. As coupled with the generator, an encoder can potentially learn to produce latent high-semantic features from the initial data distribution. This can bring an important advancement in solving inverse problems (e.g., image inpainting and noise removal) and downstream discrimination tasks. Two of the most seminal contributions on combining the power of GANs with Autoencoders, are BiGAN \cite{paper7} and AAE \cite{paper10}. 

Although we use the same architectural components (i.e., Encoder, Decoder and Discriminator), our framework differs from the previous mentioned ones in several glaring aspects. First, our discriminator operates in the data space. In contrast, the critic of AAE processes samples from the latent space, while BiGAN framework concatenates a sample from the data space with its projection in the embedded space, before feeding it to the discriminator. Second, in BiGAN, the encoder and decoder can not directly communicate with each other. Therefore, the objective function of this approach does not have an explicit reconstruction cost. However, AAE and our model explicitly minimize the cycle cost. Unlike AAE, where both the encoder and decoder are trained to perform reconstruction, our encoder weights are frozen, while optimizing with respect to the cycle loss, in order to avoid drifting the features learned using the clustering loss. Moreover, in BiGAN and AAE, the encoder and decoder are trained jointly in competition with the discriminator network. However, in our framework, each network is trained separately. Furthermore, AAE and BiGAN are standard generative models. So, in order to allow sampling, they enforce the aggregated posterior to match an arbitrary prior. 
However, in our case, we do not impose any hypothetical prior distribution and the adversarial training strategy is introduced to tackle problems related to embedded clustering, that is, Features Randomness and \textit{Feature Drift}.

\section{Trade-Off Between Feature Randomness and Feature Drift}
In this section, we propose a mathematical formalism to characterize \textit{Feature Randomness} and \textit{Feature Drift}. We explain the identified problems and we shed light on the trade-off between them.

\subsection{Feature Randomness}
For a pseudo-supervised loss, the used labels are predicted based on a presumptive similarity metric. It then follows that part of the pseudo-labels mismatch the real ones. Zhang et al. \cite{paper66} showed empirically that standard deep neural networks can easily and perfectly fit completely random labels without any considerable time overhead, using the same hyperparameters and architecture as used for training with correct labels. This result suggests that a neural network has sufficient capacity to memorize the whole dataset even when there is little or no correlation between the training samples and their corresponding labels.

We call \textit{Feature Randomness} the training of a neural network using pseudo-labels. Put it differently, \textit{Feature Randomness} takes place when a significant portion of the true labels are substituted by random ones. At every iteration of a neural network optimization process, \textit{Feature Randomness} can be characterized by $\Delta_{FR}$ (\ref{eq:delta_FR}). $\Delta_{FR}$ is the cosine of the angle between the gradient of the unsupervised loss and the gradient of the real supervised loss w.r.t the network parameters $w$. $y_{true}$ denotes the true labels (100\% correct) and $y_{pseudo}$ denotes the pseudo-labels (partially correct).


\begin{equation} \label{eq:delta_FR}
    \Delta_{FR}= cos(\widehat{\frac{\partial L(x, y_{true}, w)}{\partial w}, \frac{\partial L(x, y_{pseudo}, w)}{\partial w}}).
\end{equation}

\begin{figure}[ht]
\vskip 0.2in
\begin{center}
\centerline{\includegraphics[width=300pt, height=80pt]{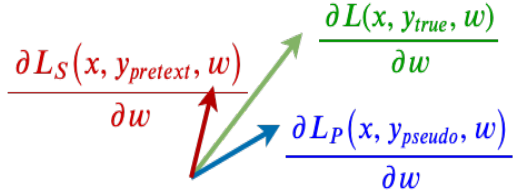}}
\caption{Adjusting the gradient of the pseudo-supervised loss by the gradient of a self-supervised loss to better approximate the gradient of the true supervised loss.}
\label{fig:fr_Ls}
\end{center}
\end{figure}

Training with pseudo-labels deteriorates the generalization capacity of a neural network. In fact, it enforces learning features that emphasize similarities between uncorrelated data points. In order to reduce the harm of \textit{Feature Randomness}, a possible solution consists of adjusting the gradient of the pseudo-supervised loss $L_{p}$ by another vector. The gradient of $L_{s}$ is a candidate to be that vector for several reasons. First, it is well known that minimizing $L_{s}$ generates reasonable general-purpose features. Second, the self-acquired labels for $L_{s}$ are 100\% correct. Hence, minimizing $L_{s}$ does not contribute to \textit{Feature Randomness}. Besides, self-supervision can be used to integrate relevant prior-knowledge. In Figure \ref{fig:fr_Ls}, we illustrate the role of the self-supervised objective function in adjusting the gradient of the pseudo-supervised one. $y_{pretext}$ denotes the pretext-labels.
 

\subsection{Feature Drift}
In the context of multi-objective optimization, two objective functions are said to be conflicting, if optimizing one of them in value degrade the other one. In such a case, the optimum should be computed, while taking into consideration the trade-offs between the competing objective functions. A solution is called non-dominated when there are multiple optima that jointly optimize the objective functions. These optima are considered equally valid in the absence of extra subjective information.

\begin{figure}[ht]
\vskip 0.2in 
\centering
    \subfigure[Strong competition between $\vec{A_{1}}$ and $\vec{B_{1}}$.]{\includegraphics[width=200pt, height=120pt]{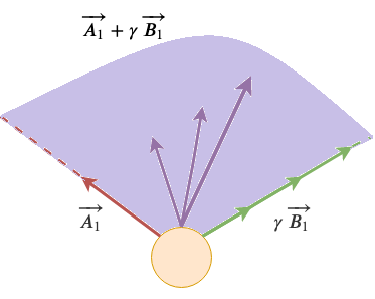}}
    \subfigure[Weak competition between $\vec{A_{2}}$ and $\vec{B_{2}}$.]{\includegraphics[width=180pt, height=120pt]{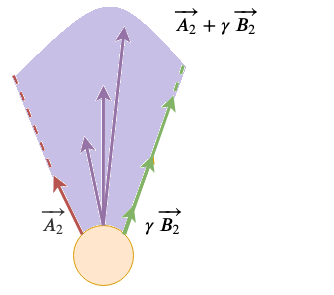}}
    \caption{Comparing the impact of strong and weak competition in reaching a target position.}
    \label{fig:competition}
\end{figure}

In the case of deep learning, we call \textit{Feature Drift} the optimization of a neural network's loss function whose secondary component (regularizer) \textit{strongly} competes with the main one.
This phenomenon can lead to a failure of the global learning process. The features learned based on the main cost function can be easily drifted by updating with respect to the secondary loss. To better understand this problem, Figure \ref{fig:competition} shows a simplistic illustration of \textit{Feature Drift}. In Figure \ref{fig:competition}.a, a linear combination of two strongly competing vectors $\overrightarrow{A_{1}}$ and $\overrightarrow{B_{1}}$ is pulling the ball. In Figure \ref{fig:competition}.b, the ball is pulled by another couple of forces $\overrightarrow{A_{2}}$ and $\overrightarrow{B_{2}}$, which are less conflicting. In both cases, the pulling forces are adjusted using a balancing positive coefficient $\gamma$. So, the resultant vector is equal to $\overrightarrow{A_{1}} + \gamma \: \overrightarrow{B_{1}}$ for the first case and  $\overrightarrow{A_{2}} + \gamma \: \overrightarrow{B_{2}}$ for the second case. In this example, we consider that the pulling forces $\overrightarrow{A_{1}}$, $\overrightarrow{A_{2}}$, $\overrightarrow{B_{1}}$, and $\overrightarrow{B_{2}}$ are constants and the $\gamma$ coefficient is variable. After applying the resultant vector, the object is supposed to reach a position $P(\gamma)$. In both figures, the colored area (delimited by the competing vectors) represents the field of possible positions after applying the resultant vector. The target solution lies within the green field. It is reasonable to predict that the smaller the area of the field, the easier to reach the target solution.
For two strongly competing vectors, we observe that the variation of $\gamma$ dramatically affects the reached position. Therefore,  the choice of $\gamma$ is quite critical for this case. Whereas, for the second case, where the competition is less ardent, the variation of $\gamma$ has a less important impact on the reached position. Hence, the choice of $\gamma$ is less crucial than in the first case. We conclude that the importance of a balancing coefficient depends on the level of competition. Most importantly, we observe that the level of competition can be assessed by the cosine of the angle between the two competing vectors. Hence, at every training iteration, we can characterize \textit{Feature Drift} as following:

\begin{equation} \label{eq:delta_FD}
    \Delta_{FD}= cos(\widehat{\frac{\partial L_{p}(x, y_{pseudo}, w)}{\partial w}, \frac{\partial L_{s}(x, y_{pretext}, w)}{\partial w}}).
\end{equation}

Where $\frac{\partial L_{p}}{\partial w}$ and $\frac{\partial L_{s}}{\partial w}$ are the gradient of the pseudo-supervised loss and the gradient of the self-supervised loss, respectively. When the strongly contending vectors are not balanced meticulously, the desired solution would not be reached even after multiple iterations. Besides, it is of great importance to make unsupervised learning methods less reliant on unpredictable and dataset-specific parameters. The example presented by Figure \ref{fig:competition} was selected for its simplicity, since it is difficult to visualize the gradient vectors in a high-dimensional space.

Several modern deep clustering models \cite{paper28, paper29, paper36, paper35} jointly perform reconstruction and embedded clustering. In this work, we show empirically that combining them leads to a very strong competition between their gradients. Under specific conditions, we provide a mathematical proof of this hypothesis. For this reason, we consider the problem of clustering a dataset $DS = \left \{ x_{i} \in \mathbb{R}^{n} \right \}_{i=1}^{N}$ and X the matrix whose raw vectors are $x_{i} \in DS$. We presume that the number of clusters $K = 2$. Our operators include a linear encoder $E_{A}:\mathbb{R}^{n} \rightarrow \mathbb{R}^{d}$, which maps samples from the data space to the latent space and a linear decoder $G_{B}$ performing an inverse mapping. The matrices $A$ and $B$ hold the learnable parameters of the encoder and decoder, respectively. We define the vector $z_{i} = E_{A}(x_{i}) = A \; . \; x_{i}$  as the projection of the data point $x_{i}$ in the latent space and $\hat{x_{i}} = G_{B}(z_{i}) = B \; . \; z_{i}$ is the reconstructed representation of $x_{i}$. We denote $\hat{z_{i}} =  A \; . \; B \; . \; z_{i}$. Furthermore, we constrain $A$ to the set of semi-orthogonal matrices. Thus, $A^{T}\; . \; A = I_d$, where $I_d$ is the identity matrix. Each cluster is associated with a centroid $\mu_{j}=\frac{1}{N_{j}}\sum_{i \in C_{j}} z_{i}$ in the embedded space $\mathbb{R}^{d}$, where $j \in [|1,K|]$,  $C_{j}$ represents the cluster j and $N_{j}$ is the number of points in $C_{j}$. The center of the embedded points is denoted by $\bar{z}=\frac{1}{N}\sum_{i=1}^{N} z_{i}$. We define, $L_{r}=\sum_{i=1}^{N} (x_{i} - \hat{x}_{i})^{2}$, as the reconstruction loss, and $L_{k}= \sum_{j=1}^{K} \sum_{i \in C_{j}}(z_{i} - \mu_{j})^{2}$, as the k-means loss in the latent space.



Let $d(C_{l_{1}}, C_{l_{2}}) = \sum_{i \in C_{l_{1}}} \sum_{j \in C_{l_{2}}}(z_{i} - z_{j})^{2}$, be the average distance between two clusters $C_{l_{1}}$ and $C_{l_{2}}$. If $l_{1}$ is equal to $l_{2}$, this distance is called within-cluster distance, and defined by $C_{l_{1}}$ and $C_{l_{2}}$. Otherwise, it is called between-cluster distance, and defined by $C_{l_{1}}$ and $C_{l_{2}}$.


\begin{theorem}
Under the specific conditions described above, the loss function $L_{DCN} = L_{k} + \gamma L_{r}$ can be expressed as following:

\begin{equation} \label{eq:ft_dcn}
    L_{DCN}= (1+\gamma) J_{1} - \frac{1}{2} J_{2} + \gamma J_{3},
\end{equation}

where 

\begin{equation*} 
        \begin{split}
        J_{1} & \: = \:  \frac{1}{N}\;d(C_{1},C_{2})+\frac{1}{2N}\;d(C_{1},C_{1})+\frac{1}{2N}\;d(C_{2},C_{2}),\\ 
        J_{2} & \: = \: \frac{N_{1}N_{2}}{N}\;\left [ 2 \; \frac{d(C_{1},C_{2})}{N_{1}N_{2}}-\frac{d(C_{1},C_{1})}{N_{1}^{2}}-\frac{d(C_{2},C_{2})}{N_{2}^{2}}\right ],\\
        J_{3} & \: = \: \sum_{i=1}^{N}(\hat{z}_{i}-\bar{z})^{2} -2(z_{i}-\bar{z})^{T}(\hat{z}_{i}-\bar{z}).\\ 
        \end{split}
\end{equation*}

\end{theorem}

The proof of Theorem 1 is provided in Appendix A. This theorem shows the implicit competition between a typical clustering loss (k-means) and the reconstruction one. Intuitively, minimizing the clustering loss has two objectives. First, it allows to emphasize the similarities between data points within the same cluster. Second, it enables to stress the variations between data points from different clusters. However, minimizing the reconstruction loss aims to preserve all the similarities and variances between every couple of data points, whether or not they belong to the same cluster. Using Theorem 1, we can notice that minimizing the first term of $J_{2}$ leads to the maximization of between-cluster distances, which force the clusters to be separable. Added to that, minimizing the second and third terms of $J_{2}$  minimizes within-cluster variances, which pushes the clusters to be as compact as possible. However, minimizing $J_{1}$ implies minimizing both between-cluster distances and within-cluster variances.

In a general case, increasing $\gamma$ significantly causes the self-supervised loss to easily win the competition. Thus, any discriminative feature learned in the direction of the pseudo-supervised loss's gradient can be easily drifted by the gradient of the self-supervised loss. On the other side, decreasing $\gamma$ significantly leads to \textit{Feature Randomness}.

\section{Adversarial Deep Embedded Clustering}
In this section, we describe our proposed framework ADEC. Our model is designed to address \textit{Feature Randomness} and \textit{Feature Drift}. To this end, we consider the problem of clustering a dataset $ X = \left \{ x_{i} \in \mathbb{R}^{n} \right \}_{i=1}^{N}$ into $K$ clusters. Each cluster is associated with a centroid $\mu_{j}$ in the embedded space $\mathbb{R}^{d}$, where $j \in [|1,K|]$. Our operators include a deep non-linear encoder $E_{\phi }:\mathbb{R}^{n} \rightarrow \mathbb{R}^{d}$, which maps samples from the data space to the latent space and a deep non-linear decoder $G_{\theta}$ performing an inverse mapping. $\phi$ and $\theta$ represent the learnable parameters of the encoder and decoder, respectively. We define the vector $z_{i} = E_{\phi}(x_{i})$ to be the projection of a data point $x_{i}$ into the latent space and $\hat{x_{i}} = G_{\theta}(z_{i})$ is the reconstructed representation of $x_{i}$. 

\subsection{Pretraining phase}
Following state-of-the-art autoencoder-based clustering approaches, we pretrain the encoder and decoder. In the context of deep clustering, pseudo-labels are the cluster representatives (i.e., embedded centers). It is important to start the clustering phase with latent features that reflect the data distribution. Otherwise, it would be impossible to extract meaningful pseudo-labels. Training a neural network using embedded centers extracted from completely random latent representations, leads to excessive \textit{Feature Randomness} (due to the large number of unreliable pseudo-labels). It is well-known that self-supervision allows to learn reliable general-purpose features by solving a pretext task. Therefore, the pretraining phase should consist of minimizing a self-supervised loss.

\begin{figure}[ht]
\vskip 0.2in
\begin{center}
\centerline{\includegraphics[width=400pt, height=180pt]{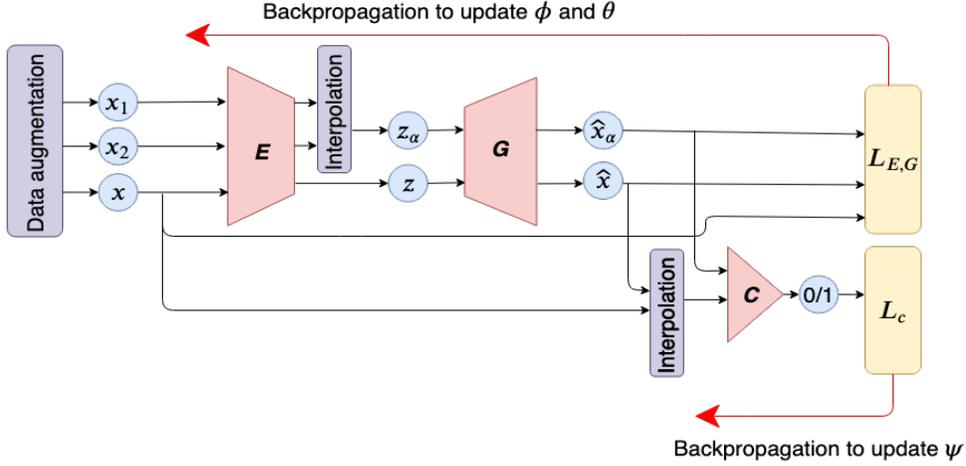}}
\caption{The pretraining phase of ADEC.}
\label{fig:pretraining}
\end{center}
\end{figure}

Previously proposed algorithms, such as, DEC and IDEC rely on a stacked denoising self-encoding strategy \cite{paper33} for initializing the training weights $\phi$ and $\theta$. In our case, we opted for pretraining the autoencoder using vanilla reconstruction loss regularized by an adversarially constrained interpolation \cite{paper5} and data augmentation (e.g., slight random rotation and translation of the input samples) \cite{paper34}. These techniques are backed up by results showing an important enhancement in learning unsupervised representations for downstream tasks \cite{paper5, paper34}. When pretraining the model, a real number $\alpha \in [0,1]$ is randomly sampled to compute $\hat{x}_{\alpha}$, such that, $\hat{x}_{\alpha} = G_{\theta}(\alpha E_{\phi}(x_{1}) + (1-\alpha) E_{\phi}(x_{2}))$ is the reconstruction of a data point interpolated from the latent representations of $x_{1}$ and $x_{2}$. The framework of ACAI \cite{paper5} simulates a game competition between two adversarial networks. The autoencoder is trained to generate interpolated points. While the critic $C_{\psi}$, which is a neural network parameterized by $\psi$, enables to regress the interpolation parameter $\alpha$ in (\ref{eq:L_C}), the autoencoder aims to fool the critic into considering the generated interpolants as real samples (i.e., outputting $\alpha=0$) in (\ref{eq:L_E_G}). The second term in (\ref{eq:L_C}) allows the critic to identify non-interpolated inputs. The coefficient $\gamma$, in (\ref{eq:L_C}), is randomly selected from $[0,1]$ at every iteration and $\lambda$, in (\ref{eq:L_E_G}), is responsible for balancing the reconstruction and the regularization. For the sake of simplification, we assume that $x$ stands for the data samples after carrying out the random transformations (rotation and translation). The full framework of our pretraining phase is illustrated in Figure \ref{fig:pretraining}. To the best of our knowledge, we are the first to propose such a pretraining strategy in the context of deep clustering.

\begin{equation} \label{eq:L_E_G}
    L_{E,G}(\phi(t), \theta(t)) =  ||x - \hat{x}||_{2}^{2} + \lambda ||C_{\psi}(\hat{x}_{\alpha})||_{2}^{2},
\end{equation}

\begin{equation} \label{eq:L_C}
    L_{C}(\psi(t)) = ||C_{\psi}(\hat{x}_{\alpha}) -\alpha||_{2}^{2} + ||C_{\psi}(\gamma x+(1-\gamma)\hat{x})||_{2}^{2}.
\end{equation}

\subsection{Clustering phase}
For this phase, on top of the pretrained encoder and decoder, we need one additional network. More precisely, we introduce a Discriminator $D_{\omega}:\mathbb{R}^{n} \rightarrow \left [ 0,1 \right ]$. Similar to the standard GAN framework, the Discriminator allows to identify real data samples from the fake ones. This network is parameterized with $\omega$. Based on our experimental results, the feature learned by minimizing the embedded clustering objective function can be easily drifted, while minimizing the reconstruction loss. To inhibit this implicit strong competition from taking place, our strategy aims at transforming within-network competition to a different between-networks one. Therefore, each network is trained independently from the other ones to avoid the drifting effect. 

Training the encoder is the main step in our framework. The clustering loss in (\ref{eq:L_E}) is inspired by DEC
\cite{paper27}. It refines the clusters by gradually stressing high confidence assignments. It is worth to note that our methodology can be applied using a different embedded clustering loss. The choice of the DEC cost can be explained by its simplicity and popularity in the deep clustering community. Unlike DEC, we add a regularization term (second part of (\ref{eq:L_E})). Our regularization allows to penalize generating embedded features, which could not be decoded into realistic data points. This constraint is verified by the discriminator, leading to rejecting discriminative features, which corrupt the clustering space. Hence, we argue that minimizing (\ref{eq:L_E}) enables to reduce \textit{Feature Randomness}.

\begin{equation} \label{eq:L_E}
    \begin{split}
    L_E(\phi(t)) & =  KL(P||Q) \: + \\
    &  \: \mathbb{E}_{x \sim p(x)}[log(1-D_{\omega}(G_{\theta}(E_{\phi}(x)))].
    \end{split}
\end{equation}

As shown by equation (\ref{eq:L_E}), our model does not require a balancing hyperparameter. Unlike IDEC, where the balancing hyperparameter is critical and hard-to-tune due to the strong trade-off, in our case, the clustering and regularization terms do not reflect any explicit competition. We would provide an experimental study on hyperparameter tuning to validate the aforesaid hypothesis. 

Unlike DEC, where the decoder is discarded straight away, this network plays a pivotal role in our case. It can be seen as a monitor. It allows to investigate the variations of the embedded representations induced by training the encoder. Hence, we argue that the decoder should be trained as well to catch-up with the encoder updates. However, training the decoder similar to IDEC would drift the discriminative features learned by the encoder. We propose to restrain the backpropagation of the reconstruction loss to the decoder layers as shown by equation (\ref{eq:L_G}). We argue that such a strategy helps in reducing \textit{Feature Drift}.

\begin{equation} \label{eq:L_G}
    L_G(\theta(t)) = \mathbb{E}_{x \sim p(x)}[\left \| x - G_{\theta}(E_{\phi}(x)) \right \|_{2}^{2}].
\end{equation}

\begin{figure}[ht]
\vskip 0.2in
\begin{center}
\centerline{\includegraphics[width=300pt, height=180pt]{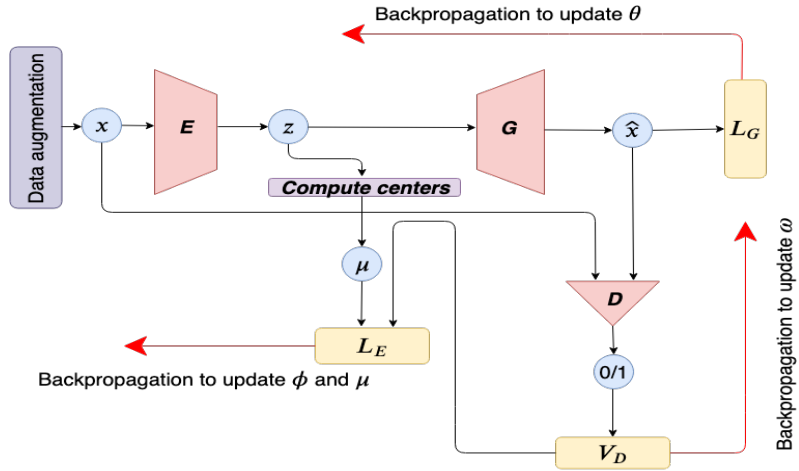}}
\caption{The clustering phase of ADEC.}
\label{fig:clustering_arc}
\end{center}
\end{figure}

Unlike DEC and IDEC, we introduce a discriminator as an additional architectural component. As exhibited by equation (\ref{eq:V_D}), the discriminator is supposed to differentiate real data points from those generated randomly. Similar to the decoder, the discriminator should be sufficiently trained before updating the encoder weights at every clustering iteration. An illustration of the clustering phase is given by Figure \ref{fig:clustering_arc}.

\begin{equation} \label{eq:V_D}
    \begin{split}
    V_D(\omega(t))  & =  \mathbb{E}_{x \sim p(x)}[log(D_{\omega}(x))] \: + \\ & \: \mathbb{E}_{x \sim p(x)}[log(1-D_{\omega}(G_{\theta}(E_{\phi}(x)))].
    \end{split}
\end{equation}

\begin{figure}[ht]
\vskip 0.2in
\begin{center}
\centerline{\includegraphics[width=\columnwidth, height=48pt]{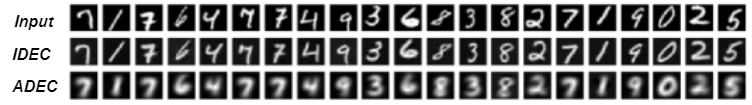}}
\caption{First row: MNIST input images; Second row: Output images from IDEC; Third row: Output images from ADEC.}
\label{fig:recons_ADEC_IDEC}
\end{center}
\end{figure}

At the end of the training process, we observe that the output images are smoother. Added to that, they do not represent a pure reconstruction anymore even if the decoder is trained for a huge number of iterations. This suggests that the encoder learned to destroy non-discriminative information. Another interesting observation is that the decoder maps images from the same class to the same blurry output image. This observation suggests that the encoder has learned to collapse within-class variances. Such characteristics of our model are inconsistent with IDEC pure reconstruction as illustrated by Figure \ref{fig:recons_ADEC_IDEC}. 

\subsection{Optimization}

ADEC has five kind of learnable parameters $\phi$, $\mu$, $\theta$, $\omega$ and $\psi$. All of them are updated using mini-batch SGD and backpropagation. $\tilde{L}_{G}$, $\tilde{V}_{D}$ and $\tilde{L}_{E}$ are, respectively, the stochastic mini-batch approximation of $L_{G}$, $V_{D}$ and $L_{E}$. The gradients are computed following Theorem \ref{theorem:gradient_z} and Theorem \ref{theorem:gradient_mu}. Refer to Appendix B and C for the proofs.

\begin{theorem} \label{theorem:gradient_z}
The gradient of the loss function $L_E(\phi(t))$ w.r.t the encoded data points $z_{i}$ is calculated by (\ref{eq:gradient_z}), where $(JG_{\theta})^{T}$ denotes the transpose Jacobian matrix of $G_{\theta}$ and $\triangledown D_{\omega}$ is the gradient of $D_{\omega}$. 
    \begin{equation} \label{eq:gradient_z}
        \begin{split}
        \frac{\partial L_E(\phi(t))}{\partial z_{i}} \: = \: & 2\sum _{j=1}^{K}(1+\left \| z_{i}-\mu_{j} \right) \|^{2})^{-1}(p_{ij}-q_{ij})(z_{i}-\mu_{j}) \: + \\
        &  \ \frac{(JG_{\theta}(z_{i}))^{T} \: .\:\triangledown D_{\omega}(G_{\theta}(z_{i}))}{1 - D_{\omega}(G_{\theta}(z_{i}))}.
        \end{split}
    \end{equation}
\end{theorem}

\begin{theorem} \label{theorem:gradient_mu}
The gradient of $L_E(\phi(t))$ w.r.t. to the cluster center $\mu_{j}$ is computed following (\ref{eq:gradient_mu}).
    \begin{equation}  \label{eq:gradient_mu}
        \begin{split}
        \frac{\partial L_E(\phi(t))}{\partial \mu_{j}} \: = \: -2 \sum _{j=1}^{K}(1+\left \| z_{i}-\mu_{j} \right) \|^{2})^{-1}(p_{ij}-q_{ij})(z_{i}-\mu_{j}).
        \end{split}
    \end{equation}
\end{theorem}

For the clustering phase, we run the optimization for $MaxIter$ batch iterations or until the clustering assignment variation between two consecutive clustering iterations is lower than $tol\%$. We found empirically that the decoder needs to be trained for a greater number of iterations compared to the other networks, otherwise it would cause instability. Thus, we alternate between training the \{Decoder $G_{\theta}$, Encoder $E_{\phi}$, Discriminator $D_{\omega}$\} for $M$ number of iterations and training the \{Decoder $G_{\theta}$\} alone also for $M$ auxiliary iterations. The target distribution $P$, which is computed based on the predicted clustering assignment distribution $Q$, constitutes the support for computing the pseudo-labels. $P$ is updated every $T$ iterations based on equations (\ref{eq:q_ij}) and (\ref{eq:p_ij}). In practice, we refrain from bringing modifications on $P$ at every single step to avoid instability. The predicted label $y_{pred}(i)$ for a data point $x_{i}$ is calculated based on the following equation: 

\begin{equation} \label{eq:_ypred}
    y_{pred}(i) = argmax_{j}(q_{ij}).   
\end{equation}

Our proposed algorithm is summarized in Algorithm \ref{alg:algorithm1}. 

\begin{algorithm}
\caption{Minibatch stochastic gradient descent training of Adversarial Deep Embedded Clustering.}
\label{alg:algorithm1}
\textbf{Input}: Input data: $X$, Number of clusters: $K$, Learning rate: $\vartheta$, Convergence threshold: $tol$, Maximum iterations: $MaxIter$, Auxiliary iterations: $M$, Distribution update interval: $T$.  \\
\textbf{Output}: Encoder weights: $\phi$, Decoder weights: $\theta$, Discriminator weights: $\omega$, Embedded centroids: $\mu$.
\begin{algorithmic}
\STATE Pretrain the autoencoder by minimizing (\ref{eq:L_E_G}) and (\ref{eq:L_C}).
\STATE Pretrain the discriminator by maximizing (\ref{eq:V_D}).
\STATE Initialize the embedded centroids $\mu$ using k-means.
\STATE $test \gets True, \; \;  j \gets 0$ .
\FOR{i$=$0 {\bfseries to} $MaxIter$}
    \IF {($i \: mod \: T \; == \; 0$)}
        \STATE Update $Q$ and $P$ using (\ref{eq:q_ij}) and (\ref{eq:p_ij}).
        \STATE Save last predicted labels: $y_{pred\_old} \gets y_{pred}$.
        \STATE Compute the new predicted labels $y_{pred}$ using (\ref{eq:_ypred}).
        \IF {($\frac{1}{N}\sum_{i=1}^{N}(y_{pred}\neq y_{pred\_old})<tol$)}
            \STATE End training.
        \ENDIF
    \ENDIF
    \IF {($test \; == \; True$)}
        \STATE $\theta \gets \theta - \vartheta \triangledown_{\theta} \tilde{L}_{G}$.
        \STATE $j \gets j + 1$.
        \IF {($j \; > \; M$)}
            \STATE $test \gets False, \; \; j \gets 0$.
        \ENDIF
    \ELSE
        \STATE $\phi \gets \phi - \vartheta \triangledown_{\phi}\tilde{L}_{E}$ ($\tilde{L}_{E}$ is computed using (\ref{eq:L_E})).
        \STATE $\theta \gets \theta - \vartheta \triangledown_{\theta} \tilde{L}_{G}$ ($\tilde{L}_{G}$ is computed using (\ref{eq:L_G})).
        \STATE $\omega \gets \omega + \vartheta \triangledown_{\omega} \tilde{V}_{D}$ ($\tilde{V}_{D}$ is computed using (\ref{eq:V_D})).
        \STATE $\mu \gets \mu - \vartheta \triangledown_{\mu}\tilde{L}_{E}$.
        \STATE $j \gets j + 1$.
        \IF {($j \; > \; M$)}
            \STATE $test \gets True, \; \; j \gets 0$.
        \ENDIF
    \ENDIF
\ENDFOR
\end{algorithmic}
\end{algorithm}

\section{Experiments}
An extensive experimental protocol is conducted to validate the suitability of ADEC in tackling \textit{Feature Randomness} and \textit{Feature Drift}. In order to perform this, we need to specify the scope of our experiments and the required experimental configurations. 

\subsection{Scope of experiments}
Deep Clustering models differ from each other in five substantial aspects. Each one of these aspects has been proved to have a significant impact on the clustering quality.
The first factor is the used \textit{architecture}. Some studies \cite{paper24, paper71} rely on sophisticated architectures (e.g., ResNet32, AlexNet and VGG) to cluster very large datasets. Other studies \cite{paper27, paper28} leverage fairly sized architectures to cluster sizeable datasets. Likewise, in this paper, we opted for the same architecture used by \cite{paper27, paper28, paper35, paper29}. The second factor is the integrated \textit{prior knowledge} (e.g. invariance of images' labels to small linear transformations and symmetries). For instance, previous works \cite{paper26, paper34} have proved that data augmentation based on prior knowledge leads to better clustering results. Inspired by these studies, we apply a similar data augmentation technique for image datasets.
The third factor of quality is the \textit{learning dynamics}. This was the focus of the following studies: (1) deep over-clustering \cite{paper71}, which offers two sub-heads, one for grouping the data in more clusters than required, and the second for clustering according to the ground truth number of clusters; (2) deep adaptive clustering \cite{paper23}, which clusters the easy samples first and then gradually supply the learning model with more difficult ones; and (3) clustering with a dynamic loss function \cite{paper72}, which gradually change the cost function according to the clustered samples. Unlike these papers, ADEC does not rely on any specific learning dynamics. Finally, the fourth and fifth factors consist in choosing the \textit{self-supervised} and \textit{pseudo-supervised} losses, respectively. \textit{Jabi et al.} \cite{paper73} proved that, under mild conditions, several pseudo-supervised objective functions are equivalent to each other.

All the previous deep clustering studies revolve around the five mentioned axes. The modification of any one of these factors is deemed to improve or worsen the effectiveness and efficiency of the studied deep clustering model. The following experimental protocol aims to show that the trade-off between \textit{Feature Randomness} and \textit{Feature Drift}, which was neglected by previous studies, is influential in designing deep clustering models. For this reason, our experiments should include a comparison, where all the other factors of quality are kept identical between ADEC and its baselines. 

\subsection{Experimental Settings}
All experiments are conducted on a server with 4 Intel(R) Xeon(R) CPU E5-2660 0 @ 2.20GHz, 32 GO RAM and a NVIDIA TESLA K80 GPU.

\subsubsection{Datasets}
We evaluate our approach on six benchmark datasets:

\begin{itemize}
    \item MNIST-full \cite{paper38}: a 10 classes database of $70,000$ grayscale handwritten digit images of size $28\times28$ each. 
    \item MNIST-test: a subset of $10,000$ images of the MNIST-full dataset. 
    \item USPS \cite{paper39}: a 10 classes  database of $9,298$ grayscale digit images of size $16\times16$ each. 
    \item Fashion-MNIST \cite{paper40}: a 10 classes database of $70,000$ grayscale images of size $28\times28$ each.
    \item REUTERS-10K \cite{paper69}: a 4 classes database (corporate/industrial, government/social, markets and economics) of $10,000$ articles. The $2,000$ most frequent words in all articles are selected. Then, for each article, we compute the TF-IDF features using the selected dictionary.
    \item Mice Protein \cite{paper70}: an 8 classes database of $1,080$ mice samples. The features of this database consists of the expression levels of 77 proteins. 
\end{itemize}

All datasets are normalized before being fed to the clustering models, thereby the norm of each data point $\frac{1}{n}\left \| x_{i} \right \|_{2}^{2}$ is approximately equal to 1. For fully-connected models, we flatten the input data if its dimension is greater than one.

\subsubsection{Baselines}
In order to show the effectiveness of the proposed model, ADEC is compared against classical clustering algorithms, subspace clustering algorithms, manifold clustering algorithms, and state-of-the-art deep clustering algorithms. The classical clustering baselines include k-means \cite{paper14}, Gaussian mixture models (GMM) \cite{paper47}, Least Squares Non-negative Matrix Factorization (LSNMF) \cite{paper75} and agglomerative clustering (AC) \cite{paper74}. The subspace clustering methods include Scalable Sparse Subspace Clustering by Orthogonal Matching Pursuit (SSC-OMP) \cite{paper78} and Scalable Elastic Net Subspace Clustering (EnSC) \cite{paper79}. The other subspace clustering baselines are not efficient enough to deal with 70,000 samples and therefore they are left out. The manifold clustering approaches include normalized-cut spectral clustering (SC) \cite{paper76} and Kernel (RBF) k-means \cite{paper77}. Finally, the deep clustering algorithms include DeepCluster \cite{paper24}, JULE \cite{paper22}, SR-k-means\cite{paper73}, DEC \cite{paper27}, IDEC \cite{paper28}, DCN \cite{paper29}, VaDE \cite{paper35} and DEPICT \cite{paper36}. Our baselines also cover clustering the embedded data of an autoencoder using k-means and FINCH \cite{paper80} denoted, respectively, by (AE+k-means) and (AE+FINCH). As a side note, all the fully-connected baselines share the same architecture with ADEC. 

\subsubsection{Evaluation Metrics}
We adopt the metrics ACC \cite{paper41}, NMI \cite{paper42}, $\Delta_{FR}$ and $\Delta_{FD}$ for assessing the clustering quality. The first two metrics are widely used to compare deep clustering methods. The third and fourth metrics are among the contributions of this work. ACC and NMI lie within the range $[0, 1]$ and $\Delta_{FR}$ and $\Delta_{FD}$ lie within the range $[-1, 1]$. Higher values are better. As shown by (\ref{eq:acc}) and (\ref{eq:nmi}), ACC and NMI depend on $y_{pred}$ and $y_{true}$, where $y_{pred}$ is a vector representing the predicted labels and $y_{true}$ is the ground-truth labels vector.

\begin{equation} \label{eq:acc}
    ACC = max_{T}( \frac{\sum_{i=1}^{N} 1 \left \{ y_{true}(i)=T(y_{pred}(i)) \right \} }{N}).
\end{equation}

$T$ is selected from the set of all possible permutations mapping the predicted clusters to the ground-truth categories. The best matching can be found using the Hungarian Algorithm \cite{paper43}.

\begin{equation} \label{eq:nmi}
    NMI(y_{pred}, \;y_{true}) = \frac{I(y_{true}, \; y_{pred})}{\frac{1}{2}[H(y_{true}) + H(y_{pred})]}.   
\end{equation}

$H$ denotes the entropy and $I$ is the mutual information.

\subsubsection{Implementation}
The encoder has eight layers of dimensions $n$ - 500 - 500 - 2000 - 10. Apart from the bottleneck layer, all the other ones are activated by ReLu \cite{paper44}. The decoder is an inverse mapping of the encoding layers 10 - 2000 - 500 - 500 - $n$ with ReLu activations except for the last layer. We pretrain the autoencoder in competition with a critic for $13 \times 10^{4}$ iterations to perform data reconstruction constrained by an adversarially constrained interpolation. The learning weights are optimized using Adam \cite{paper45} with a learning rate equal to $0.0001$. $\beta_{1}$, $\beta_{2}$, and $\epsilon$ (i.e., hyperparameters specific to Adam) have the respective values $0.9$, $0.999$, and $10^{-8}$. According to ACAI \cite{paper5} paper, $\lambda$ and $\alpha$ are set equal to $0.5$ and $1$, respectively. For the clustering stage, the encoder, decoder, and discriminator are trained alternatively for $MaxIter = 10^{5}$. The training is stopped before reaching the final iteration if the convergence criterion is met. This criterion is parameterized by a threshold $tol = 0.001$.  We update the encoder, decoder and discriminator weights using SGD with a learning rate $\vartheta = 0.001$ and momentum $0.9$. All backpropagation updates are performed on random batches of size 256 for both stages (i.e., pretraining and clustering). ADEC is implemented using Python and Tensorflow \cite{paper46}.
 
\subsection{Results}
Our experimental protocol has  three parts. In the first part, our model is compared with state-of-the-art clustering algorithms. In the second part, we analyse the ability of our model to tackle \textit{Feature Randomness} and \textit{Feature Drift}. In the last part, some qualitative results are exhibited. Before showing our results, we establish some useful notations. In all the following experiments: {-} indicates OUT\_OF\_MEMORY, $\diamond$ denotes the unsuitability of the algorithm to process one-dimensional data, $\ddagger$ indicates that the pretraining phase does not support Data Transform and Adversarially Constrained Interpolation, $\dagger$ indicates that the pretraining phase does not support Data Transform, * indicates that the evaluated methods share the same pretraining weights, the same architecture, the same learning dynamics and the same clustering loss with ADEC.

\subsubsection{Comparing state-of-the art approaches}

Table \ref{table:ACC_NMI_comparison} illustrates the evaluation of several clustering approaches, including our proposed method, in terms of ACC and NMI. All baselines methods are tuned according to their default settings. First of all, we observe that state-of-the art subspace clustering algorithms, such as, SSC-OMP and EnSC are generally not suitable for clustering datasets with semantic similarities (e.g., images, text, sounds). In fact, subspace clustering presumes the data to lie in a union of low-dimensional linear subspaces. However, this assumption does not hold for datasets with clusters lieing near non-linearly shaped manifolds \cite{paper81}. Secondly, we observe that the manifold clustering approaches have better ACC and NMI values than the classical approaches on some datasets. In fact, for the manifold category, the selection of the non-linear transform is largely empirical. Particularly, no kernel space is sufficiently well-suited to effectively cluster any dataset. Thirdly, in most cases, we can see that deep clustering  models outperform all the other approaches by a huge margin. This observation confirms the suitability of deep clustering when it comes to clustering high-dimensional datasets. Finally, comparing among the deep clustering approaches, we can observe that our method provides the best results on every dataset. In terms of ACC and NMI, ADEC outperforms its state-of-the-art counterpart DEPICT by 2\% and 5\%, respectively. Worthy of note that DEPICT is the convolutional version of DEC with some minor modifications. In order to understand the outperformance of our approach, we need to conduct further experiments.

\begin{table}
  \caption{Comparison of the clustering performances in terms of ACC and NMI. The different clustering categories are separated by double horizontal lines. Best method in bold, second best emphasized.}
  \vskip 0.15in
  \begin{center}
  \begin{small}
  \begin{tabular}{|p{2.45cm}|c|c|c|c|c|c|c|c|c|c|c|c|}
    \hline
    {\textbf{Method}} & \multicolumn{2}{c|}{\textbf{MNIST-full}} & \multicolumn{2}{c|}{\textbf{MNIST-test}} & \multicolumn{2}{c|}{\textbf{USPS}} & \multicolumn{2}{c|}{\textbf{Fashion-MNIST}} & \multicolumn{2}{c|}{\textbf{REUTERS-10K}}& \multicolumn{2}{c|}{\textbf{Mice Protein}}\\
    \cline{2-13}
    & \textbf{ACC} & \textbf{NMI} & \textbf{ACC} & \textbf{NMI} & \textbf{ACC} & \textbf{NMI} & \textbf{ACC} & \textbf{NMI} & \textbf{ACC} & \textbf{NMI} & \textbf{ACC} & \textbf{NMI} \\ \hline
    \textbf{k-means} & 0.532 & 0.500 & 0.546 & 0.501 & 0.668 & 0.627 & 0.474 & 0.512 & 0.522 & 0.313 & 0.342 & 0.252 \\ \hline
    \textbf{GMM} & 0.433 & 0.366 & 0.540 & 0.493 & 0.551 & 0.530 & 0.556 & 0.557 & 0.402 & 0.375 &  0.139 & 1.00 \\ \hline
    \textbf{LSNMF} & 0.540 & 0.455 & 0.550 & 0.463  & 0.575 & 0.551 & 0.549 & 0.523 & 0.596 & 0.361 &  \textit{0.497} & \textit{0.506} \\ \hline 
    \textbf{AC} & 0.621 & 0.682 & 0.695 & 0.711 & 0.683 & 0.725 & 0.500 & 0.564 & 0.526 & 0.365 &  0.294 & 0.211 \\ \hline \hline
    \textbf{SSC-OMP} & 0.309 & 0.315 & 0.413 & 0.450 & 0.477 & 0.503 & 0.100 & 0.007 & 0.402 & 0.008 & 0.152 & 0.078 \\ \hline
    \textbf{EnSC} & 0.111 & 0.014 & 0.603 & 0.591 & 0.610 & 0.684 & \textbf{0.629} & \textit{0.636} & 0.401 & 0.014 & 0.434 & 0.347 \\ \hline \hline
    \textbf{SC} & 0.656 & 0.731 & 0.660 & 0.704 & 0.649 & 0.794 & 0.508 & 0.575 & 0.402 & 0.375 &  0.298 & 0.268 \\ \hline
    \textbf{RBF k-means} & {-} & {-} & 0.560 & 0.523 & 0.629 & 0.631 & {-} & {-} & 0.499 & 0.288 &  0.363 & 0.269 \\ \hline \hline
    \textbf{AE + k-means} & 0.807 & 0.730 & 0.702 & 0.617 & 0.720 & 0.698 & 0.585 & 0.614 & 0.695 & 0.475 & 0.238  & 0.131\\ \hline
    \textbf{AE + FINCH} & {-} & {-} & 0.709 & 0.754 & 0.704 & 0.788 & {-} & {-} & 0.241 & 0.414 & 0.157 & 0.083 \\ \hline 
    \textbf{DeepCluster} & 0.797 & 0.661 & 0.854 & 0.713 & 0.562 & 0.540 & 0.542  & 0.510 & $\diamond$ & $\diamond$ & $\diamond$ & $\diamond$ \\ \hline
    \textbf{DCN} & 0.830 & 0.810 & 0.802 & 0.786 & 0.688 & 0.683  & 0.501 & 0.558 & 0.422 & 0.109 & 0.197 & 0.051 \\ \hline 
    \textbf{DEC} & 0.863 & 0.834 & 0.856 & 0.830 & 0.762 & 0.767 & 0.518 & 0.546 &  \textit{0.814} & \textit{0.598} & 0.184 & 0.026 \\ \hline
    \textbf{IDEC} & 0.881 & 0.867 & 0.846 & 0.802 & 0.761 & 0.785 & 0.529 & 0.557 & 0.790 & 0.550 & 0.196 & 0.037 \\ \hline
    \textbf{SR-k-means} & 0.939 & 0.866 & 0.863 & 0.873 & 0.901 & 0.912 & 0.507 & 0.548 & $\diamond$ & $\diamond$ & $\diamond$ & $\diamond$ \\ \hline
    \textbf{VaDE} & 0.945 & 0.876 & 0.287 & 0.287 & 0.566 & 0.512 & 0.578 & 0.630 & 0.793 & 0.521 & 0.139 & 1.00 \\ \hline
    \textbf{JULE} & 0.964 & 0.913 & 0.961 & \textit{0.915} & \textit{0.950} & \textit{0.913} & 0.563 & 0.608 & $\diamond$ & $\diamond$ & $\diamond$ & $\diamond$  \\ \hline
    \textbf{DEPICT} & \textit{0.965} & \textit{0.917} & \textit{0.963} & \textit{0.915} & 0.899 & 0.906 &  0.392 & 0.392 & $\diamond$ & $\diamond$ & $\diamond$ & $\diamond$ \\ \hline
    \textbf{ADEC} & \textbf{0.986} & \textbf{0.961} & \textbf{0.985} & \textbf{0.957} & \textbf{0.981} & \textbf{0.948} & \textit{0.586} & \textbf{0.662} & \textbf{0.821 $\ddagger$} & \textbf{0.605 $\ddagger$} &  \textbf{0.500 $\dagger$} & \textbf{0.604 $\dagger$} \\ \hline 
  \end{tabular}
  \label{table:ACC_NMI_comparison}
  \end{small}
  \end{center}
  \vskip -0.1in
\end{table}

\begin{table}
  \caption{Comparison of the clustering performances of DEC*, IDEC*, and ADEC in terms of ACC and NMI. Best method in bold, second best emphasized.}
  \vskip 0.15in
  \begin{center}
  \begin{small}
  \begin{tabular}{|p{2.45cm}|c|c|c|c|c|c|c|c|c|c|c|c|}
    \hline
    {\textbf{Method}} & \multicolumn{2}{c|}{\textbf{MNIST-full}} & \multicolumn{2}{c|}{\textbf{MNIST-test}} & \multicolumn{2}{c|}{\textbf{USPS}} & \multicolumn{2}{c|}{\textbf{Fashion-MNIST}} & \multicolumn{2}{c|}{\textbf{REUTERS-10K}}& \multicolumn{2}{c|}{\textbf{Mice Protein}}\\
    \cline{2-13}
    & \textbf{ACC} & \textbf{NMI} & \textbf{ACC} & \textbf{NMI} & \textbf{ACC} & \textbf{NMI} & \textbf{ACC} & \textbf{NMI} & \textbf{ACC} & \textbf{NMI} & \textbf{ACC} & \textbf{NMI} \\ \hline
    \textbf{DEC*} & 0.971 & 0.929 & 0.968 & 0.920 & 0.963 & 0.910 & \textit{0.575} & 0.589 & \textit{0.814} $\ddagger$ &\textit{0.598} $\ddagger$ &  \textit{0.267} $\dagger$ & \textit{0.158} $\dagger$\\ \hline
    \textbf{IDEC*} & \textit{0.982} & \textit{0.952} & \textit{0.978} & \textit{0.944} & \textit{0.980} & \textit{0.946} & \textit{0.575} & \textit{0.631} & 0.790 $\ddagger$ & 0.550 $\ddagger$ & 0.188 $\dagger$ & 0.033 $\dagger$ \\ \hline
    \textbf{ADEC} & \textbf{0.986} & \textbf{0.961} & \textbf{0.985} & \textbf{0.957} & \textbf{0.981} & \textbf{0.948} & \textbf{0.586} & \textbf{0.662} & \textbf{0.821 $\ddagger$} & \textbf{0.605 $\ddagger$} &  \textbf{0.500 $\dagger$} & \textbf{0.604 $\dagger$} \\ \hline 
  \end{tabular}
  \label{table:fair_comparison}
  \end{small}
  \end{center}
  \vskip -0.1in
\end{table}

For a fair comparison with state-of-the-art deep clustering approaches, the baselines need to be reimplemented in a way to neutralize factors, which are out of this article scope. The new reimplemented models share the same deep clustering factors (i.e., architecture, learning dynamics, integrated prior knowledge and clustering loss) with ADEC. From Table \ref{table:ACC_NMI_comparison} and Table \ref{table:fair_comparison}, we can notice a considerable improvement in terms of ACC and NMI for the modified version of DEC and IDEC, comparatively to the original ones. More specifically, DEC* outperforms vanilla DEC by a huge margin. Similarly, IDEC* surpasses its standard counterpart significantly. The huge gap between the ordinary pretrained models (i.e., based on a simple reconstruction) and the modified ones, demonstrates the effectiveness of combining Adversarially Constrained Interpolation and Data Transformation, as a pretraining strategy. Furthermore, as we can see from Table \ref{table:fair_comparison}, ADEC* outperforms DEC* and IDEC*. This result suggests that ADEC offers a better trade-off between \textit{Feature Randomness} and \textit{Feature Drift}. This hypothesis will be further supported in the subsequent experiments.

In Table \ref{table:exec_time}, we report the execution time of different deep clustering methods. The comparison is limited to deep clustering models. Henceforth, we exclude all the other clustering categories due to their less competitive results as just demonstrated by the previous comparison in Table \ref{table:ACC_NMI_comparison}. As we can see in Table \ref{table:exec_time}, the run-time of ADEC is significantly higher than the execution times of DEC, IDEC, DCN and DeepCluster on all datasets. As it stands, these methods are more efficient than ADEC. However, we can also observe that the execution time of our method is on par with the execution times of DEPICT, SR-k-means and JULE. Interestingly, our algorithm is way faster than VaDE on all datasets. 

\begin{table}
  \caption{Comparison of the execution times (in seconds) of different deep clustering approaches.}
  \vskip 0.15in
  \begin{center}
  \begin{small}
  \begin{tabular}{|p{2.2cm}|c|c|c|c|c|c|}
    \hline
    {\textbf{Method}} & {\textbf{MNIST-full}} & {\textbf{MNIST-test}} & {\textbf{USPS}} &{\textbf{Fashion-MNIST}} & {\textbf{REUTERS-10K}}& {\textbf{Mice Protein}} \\ \hline
    \textbf{DeepCluster} & 1,375 & 74 & 64 & 1,250 & {-} & {-}\\ \hline
    \textbf{DCN} & 640 & 55 & 49 & 732 & 279 & 40 \\ \hline
    \textbf{DEC} & 693 & 58 & 53 & 2,384 & 105 & 22\\ \hline
    \textbf{IDEC} & 890 & 349 & 110 & 857 & 97 & 150\\ \hline
    \textbf{SR-k-means} & 14,872 & 1,657 & 1,655 & 4,551 & {-} & {-}\\ \hline
    \textbf{VaDE} & 123,000 & 15,000 & 13,000 & 120,000 & 105 & 15\\ \hline
    \textbf{JULE} & 12,500 & 3,247 & 2,540 & 13,100 & {-} & {-}\\ \hline
    \textbf{DEPICT} & 9,561 & 2,320 & 1,778 & 8,581 & {-} & {-} \\ \hline
    \textbf{ADEC} & 10,735 & 10,013 & 8,445 & 10,502 & 669 $\ddagger$ & 1,047 $\dagger$\\ \hline 
  \end{tabular}
  \label{table:exec_time}
  \end{small}
  \end{center}
  \vskip -0.1in
\end{table}

For fairness of comparison and in order to better assess the efficiency of our method, we run our algorithm against the modified versions of DEC and IDEC (same as the previous experiment). Based on Table \ref{table:exec_time} and Table \ref{table:fair_exec_time}, we can see that the run-time of DEC* and IDEC* are significantly higher than the execution times of vanilla DEC and vanilla IDEC, respectively. Therefore, we can conclude that DEC* and IDEC* are less efficient than DEC and IDEC, respectively. This conclusion can be explained by the long pretraining phase. Hence, the gain achieved by pretraining with an Adversarially Constrained Interpolation comes at the cost of a higher execution time. Another observation, DEC* and IDEC* are slightly faster than ADEC*. This is expected and it can be imputed to the adversarial training of our algorithm.

\begin{table}
  \caption{Comparison of the execution times (in seconds) of DEC*, IDEC* and ADEC.}
  \vskip 0.15in
  \begin{center}
  \begin{small}
  \begin{tabular}{|p{2.2cm}|c|c|c|c|c|c|}
    \hline
    {\textbf{Method}} & {\textbf{MNIST-full}} & {\textbf{MNIST-test}} & {\textbf{USPS}} &{\textbf{Fashion-MNIST}} & {\textbf{REUTERS-10K}}& {\textbf{Mice Protein}} \\ \hline
    \textbf{DEC*} & 9,667 & 9,092 & 7,692 & 10,840 & 53 $\ddagger$ & 639 $\dagger$\\ \hline
    \textbf{IDEC*} & 9,556 & 9,160 & 7,693 & 9,623 & 55 $\ddagger$ & 646 $\dagger$\\ \hline
    \textbf{ADEC} & 10,735 & 10,013 & 8,445 & 10,502 & 669 $\ddagger$ & 1,047 $\dagger$\\ \hline 
  \end{tabular}
  \label{table:fair_exec_time}
  \end{small}
  \end{center}
  \vskip -0.1in
\end{table}

\subsubsection{Feature Randomness and Feature Drift}
\label{Sec:Feature Randomness and Feature Drift}

In this section, our conducted experiments aim to show the ability of ADEC to reach a better trade-off between \textit{Feature Randomness} and \textit{Feature Drift}. Therefore, we perform an ablation of our adversarial mechanism. Instead of this mechanism, we regularize the clustering loss with vanilla reconstruction. The obtained model is IDEC*. Then, we compare ADEC with IDEC* in terms of $\Delta_{FR}$ and $\Delta_{FD}$. As mentioned earlier, both models share the same optimizer, the same pretraining phase and the same embedded clustering loss. The only difference between them is the regularization technique. 

\begin{figure*}[ht]
\vskip 0.2in
\centering
    \subfigure[ADEC.]{\includegraphics[width=0.45\linewidth]{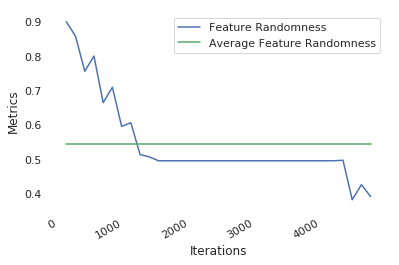}}
    \subfigure[IDEC*.]{\includegraphics[width=0.45\linewidth]{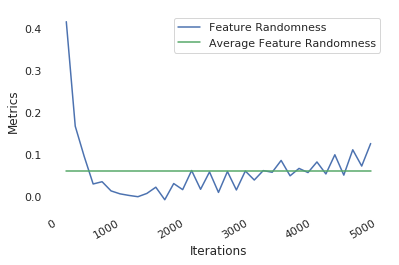}} 
    \caption{$\Delta_{FR}$ during training on MNIST.}
\label{fig:Delta_FR_MNIST}
\end{figure*}

The first experiment examines the impact of our adversarial mechanism in reducing \textit{Feature Randomness}. In this section, we show results for the MNIST dataset. Such results are representative of the general behavior of our approach and the same conclusion can be drawn on the other datasets. In Figure \ref{fig:Delta_FR_MNIST}, we draw the evolution of $\Delta_{FR}$ for ADEC and IDEC* during training on MNIST. Based on this figure, we observe that the average values of $\Delta_{FR}$ for ADEC is considerably higher than the one for IDEC*. A higher $\Delta_{FR}$ value means that the gradient of ADEC is a better approximation to the supervised gradient. Hence, this experiment confirms that our adversarial regularization is more suitable for alleviating \textit{Feature Randomness} than vanilla reconstruction. 

\begin{figure*}[ht]
\vskip 0.2in
\centering
    \subfigure[ADEC.]{\includegraphics[width=0.45\linewidth]{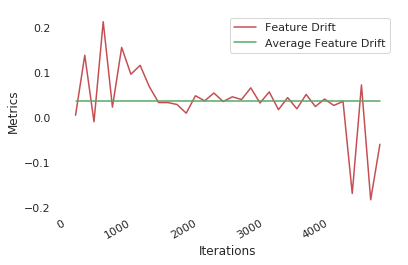}}
    \subfigure[IDEC*.]{\includegraphics[width=0.45\linewidth]{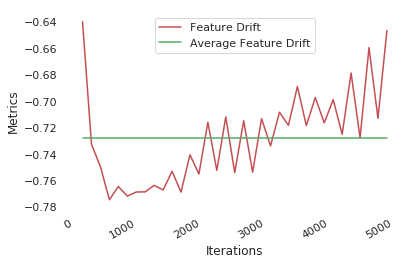}} 
    \caption{$\Delta_{FD}$ during training on MNIST.}
\label{fig:Delta_FD_MNIST}
\end{figure*}

The second experiment examined the impact of our adversarial mechanism in reducing \textit{Feature Drift}. In Figure \ref{fig:Delta_FD_MNIST}, we draw the evolution of $\Delta_{FD}$ for ADEC and IDEC* during training on MNIST. Based on this figure, we observe that the values of $\Delta_{FD}$ for IDEC* are always negative. This result confirms the strong competition between the gradient of the embedded clustering loss and the gradient of the reconstruction loss. Added to that, we observe that the average values of $\Delta_{FD}$ for ADEC is considerably higher than the one for IDEC*. A higher $\Delta_{FD}$ value indicates that the competition between the embedded clustering gradient and the reconstruction gradient is stronger than the competition between the embedded clustering gradient and the adversarial gradient. Hence, this experiment confirms that our adversarial regularization is more suitable for alleviating \textit{Feature Drift} than vanilla reconstruction. 

\begin{figure}
\centering
\begin{minipage}{.5\textwidth}
  \centering
  \includegraphics[width=200pt, height=150pt]{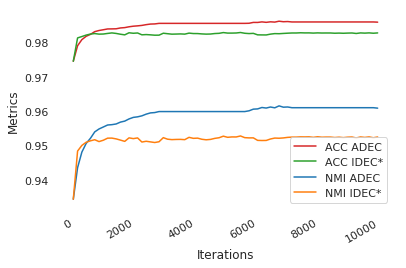}
  \caption{ACC and NMI during training on MNIST.}
  \label{fig:ACC_NMI_mnist}
\end{minipage}%
\begin{minipage}{.5\textwidth}
  \centering
  \includegraphics[width=200pt, height=150pt]{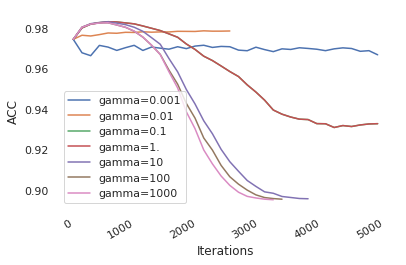}
  \caption{Sensitivity analysis for $\gamma$ during training on MNIST.}
  \label{fig:sensitivity_MNIST}
\end{minipage}
\end{figure}

\begin{figure}
\vskip 0.2in
\centering
    \subfigure[ADEC.]{\includegraphics[width=200pt, height=150pt]{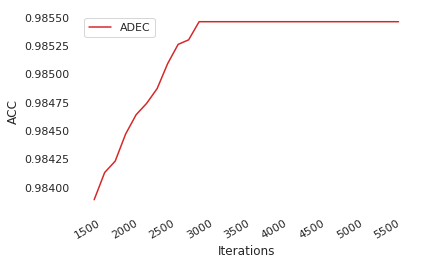}}
    \subfigure[IDEC*.]{\includegraphics[width=200pt, height=150pt]{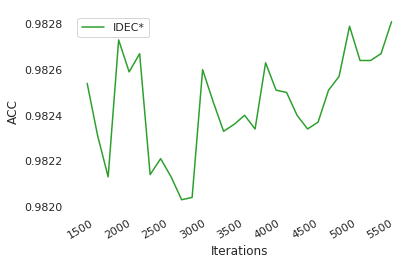}} 
    \caption{ACC during training on MNIST.}
\label{fig:ACC_MNIST}
\end{figure}

\begin{figure}
\vskip 0.2in
\centering
    \subfigure[ADEC.]{\includegraphics[width=200pt, height=150pt]{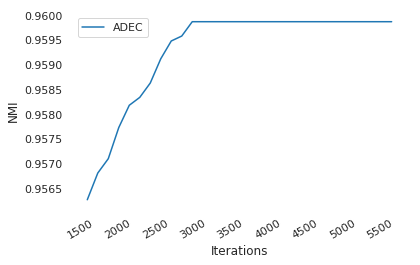}}
    \subfigure[IDEC*.]{\includegraphics[width=200pt, height=150pt]{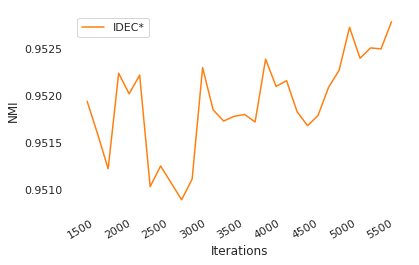}} 
    \caption{NMI during training on MNIST-test.}
\label{fig:NMI_MNIST}
\end{figure}

The third experiment examined the impact of \textit{Feature Drift} on the learning curves. In Figure \ref{fig:ACC_NMI_mnist}, we draw the learning curves of ADEC and IDEC*, in terms of ACC and NMI, during training on MNIST. Based on this figure, we observe that the learning curves of ADEC are not only above the learning curves of IDEC*, but also smoother. A zoom in to the learning curves of IDEC*, as illustrated by Figures \ref{fig:ACC_MNIST} and \ref{fig:NMI_MNIST}, shows noticeable fluctuations. However, zooming in to the learning curves of ADEC shows a smooth increase in both metrics. The observed fluctuations for IDEC* can be explained by the competition between the reconstruction and the embedded clustering.  


The fourth experiment examined the impact of \textit{Feature Drift} on the sensitivity of the balancing hyperparameter $\gamma$. In Figure \ref{fig:sensitivity_MNIST}, we draw the learning curves of IDEC*, for different values of $\gamma$, during training on MNIST. In our experiments, $\gamma$ is selected from the following set $\left\{ 10^{-3}, \: 10^{-2}, \: 10^{-1}, \: 1, \: 10, \: 10^{2}, \: 10^{3}\right\}$. Based on the obtained results, we observe that only one value of the set ($\gamma$ = 0.01) yields an acceptable learning curve. All the other values make the learning curve drop significantly. Hence, we can conclude that IDEC* is very sensitive to the choice of $\gamma$. This result can be explained by \textit{Feature Drift} (the strong competition between the gradient of the self-supervised loss and the gradient of the pseudo-supervised loss). However, in our case, ADEC does not require any balancing hyperparameter.

\subsubsection{Qualitative results}

\begin{figure*}[ht]
\vskip 0.2in

\centering
    \subfigure[MNIST-full.]{\includegraphics[width=0.32\linewidth]{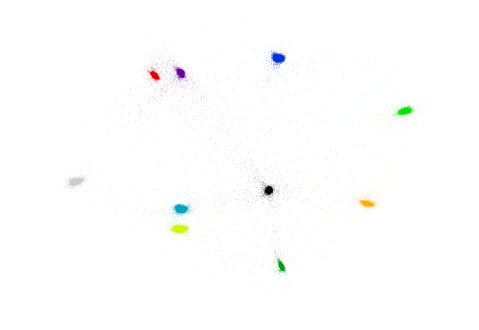}} 
    \subfigure[MNIST-test.]{\includegraphics[width=0.32\linewidth]{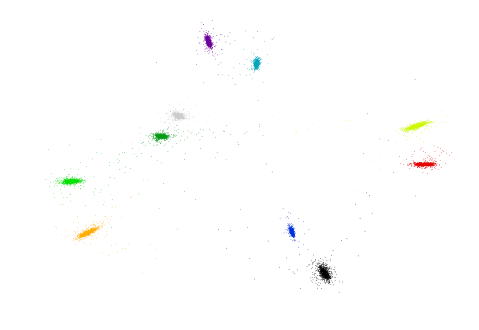}}
    \subfigure[USPS.]{\includegraphics[width=0.32\linewidth]{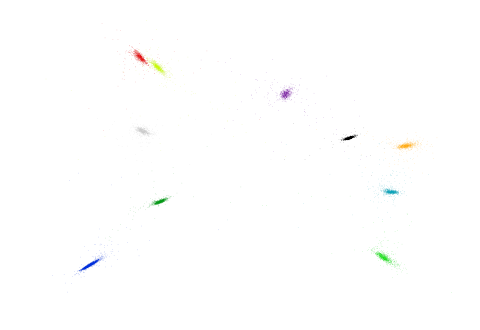}} 
    \subfigure[Fashion-MNIST.]{\includegraphics[width=0.32\linewidth]{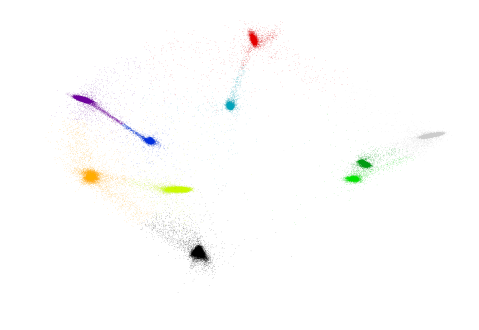}}
    \subfigure[REUTERS.]{\includegraphics[width=0.32\linewidth]{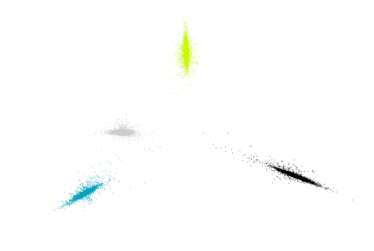}}
    \caption{2D embedding subspace visualization to show the discriminative ability of ADEC.}
\label{fig:clusetrs_embedding}
\end{figure*}

In Figure \ref{fig:clusetrs_embedding}, the discriminative ability of ADEC is illustrated by projecting the data in a 2D latent subspace for different datasets. From this figure, we can see that the projected embedded data points are grouped in well-separated clusters.

\begin{figure*}[ht]
\vskip 0.2in
\centering
    \subfigure[MNIST.]{\includegraphics[width=0.45\linewidth]{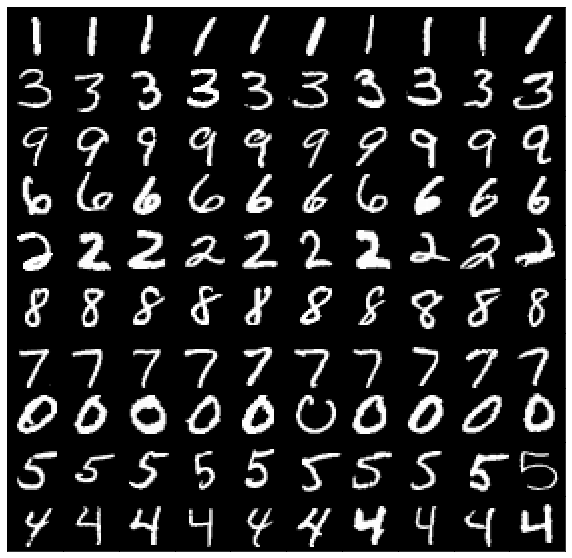}} 
    \subfigure[Fashion MNIST.]{\includegraphics[width=0.45\linewidth]{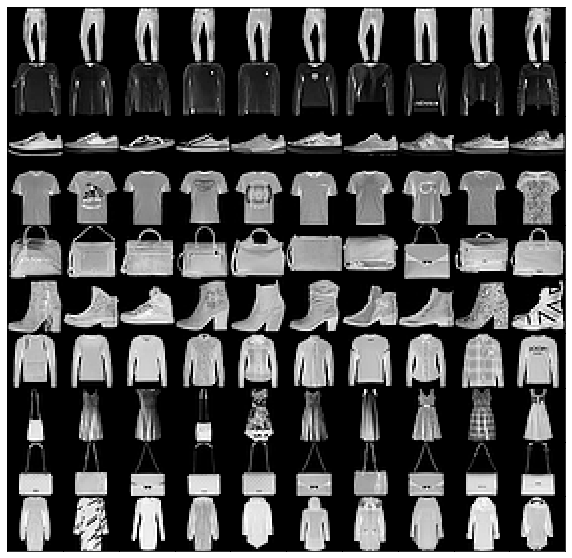}}
    \caption{Each row shows the top 10 high-confidence images from each cluster.}
\label{fig:top_cluster_images}
\end{figure*}

Figure \ref{fig:top_cluster_images} illustrates the top 10 high-confidence images from each cluster for two datasets, namely, MNIST and Fashion MNIST. In this figure, images are inserted in decreasing order from left to right according to their distance to their associated clustering centers. Every single row represents a different cluster.

\section{Conclusion}
In this article, we have proposed an Adversarial Deep Embedded Clustering algorithm. Our method enables to regularize the clustering loss in a way to alleviate \textit{Feature Randomness}. To overcome \textit{Feature Drift}, the strong clustering-reconstruction trade-off have been Eliminated. Empirical results have showed that ADEC outperforms state-of-the-art clustering methods in terms of ACC and NMI. Furthermore, our experimental studies have validated that ADEC offers a better trade-off between \textit{Feature Drift} and \textit{Feature Randomness}. For ADEC, similar to the most relevant deep clustering models, self-supervision and pseudo-supervision are combined linearly. It is very interesting to study other possible combinations and to find theoretical justifications. Besides, it is worthy to extend the scope of this work by using a more sophisticated architecture (e.g., ResNet32, AlexNet and VGG) to process higher semantic datasets.

\bibliographystyle{unsrt}


\newpage
\appendix

\section{Proof of theorem 1}

We start by computing: \\
\begin{equation*} 
        \begin{split}
        \sum_{i=1}^{N}(z_{i} - \bar{z})^{2}  & = \frac{1}{N^{2}}\sum_{i=1}^{N}\left [  \sum_{j=1}^{N}(z_{i} - z_{j})\right ]^{2},\\
        & = \frac{1}{N^{2}}\sum_{i=1}^{N} \sum_{j=1}^{N}(z_{i} - z_{j})^{2} + \frac{1}{N^{2}}\sum_{i=1}^{N} \mathop{\sum\sum}_{j\neq {j}'}(z_{i} - z_{j})^{T}(z_{i} - z_{{j}'}),\\ 
        & = \frac{d(C_{1},C_{1})}{N^{2}}  + \frac{d(C_{2},C_{2}) }{N^{2}} + \frac{2 \; d(C_{1},C_{2})}{N^{2}}  +  \frac{1}{N^{2}}\sum_{i=1}^{N}  \mathop{\sum\sum}_{j\neq {j}'}(z_{i} - z_{j})^{T}(z_{i} - z_{{j}'}).\\
        \end{split}
\end{equation*}
And since \\
\begin{equation*} 
        \begin{split}
        \sum_{i=1}^{N}  \mathop{\sum\sum}_{j\neq {j}'}(z_{i} - z_{j})^{T}(z_{i} - z_{{j}'})  & = \mathop{\sum\sum\sum}_{i\neq j,i\neq {j}',j\neq {j}'}(z_{j} - z_{i})^{2}+(z_{{j}'} - z_{i})^{2}-(z_{{j}'} - z_{j})^{2},\\
        & = (N-2) \sum_{i=1}^{N} \sum_{j=1}^{N}(z_{j} - z_{i})^{2},\\ 
        & = (N-2) \; d(C_{1},C_{1})  +  (N-2) \; d(C_{2},C_{2}) + 2 \; (N-2) \; d(C_{1},C_{2}).\\
        \end{split}
\end{equation*}
Therefore \\
\begin{equation*} 
        \begin{split}
        \sum_{i=1}^{N}(z_{i} - \bar{z})^{2}  & = \frac{1}{N}\;d(C_{1},C_{2})+\frac{1}{2N}\;d(C_{1},C_{1})+\frac{1}{2N}\;d(C_{2},C_{2}),\\
        & = J_{1}.
        \end{split}
\end{equation*}
The $L_{r}$ function can be written as: \\
\begin{equation*} 
        \begin{split}
        L_{r} & =  \sum_{i=1}^{N}(x_{i} - \hat{x}_{i})^{T}(x_{i} - \hat{x}_{i}), \\ 
        & = tr((X-BAX)(X-BAX)^{T}),\\
        & = tr(A^{T}A(X-BAX)(X-BAX)^{T}), \\ 
        & = tr(A(X-BAX)(X-BAX)^{T}A^{T}), \\ 
        & = tr((AX-ABAX)(AX-ABAX)^{T}), \\ 
        & = \sum_{i=1}^{N}(z_{i} - \hat{z}_{i})^{T}(z_{i} - \hat{z}_{i}), \\
        & = \sum_{i=1}^{N}(z_{i} - \bar{z} + \bar{z} - \hat{z}_{i})^{2}, \\
        & = \sum_{i=1}^{N}(z_{i} - \bar{z})^{2} + \sum_{i=1}^{N}(\hat{z}_{i} - \bar{z})^{2} - 2 (z_{i} - \bar{z})^{T}(\hat{z}_{i} - \bar{z}),\\
        & = \sum_{i=1}^{N}(z_{i} - \bar{z})^{2} + J_{3},\\
        & = J_{1} + J_{3}.\\
        \end{split}
\end{equation*}
According to \cite{paper65}, the $L_{k}$ function can be written as : \\
\begin{equation*} 
        \begin{split}
        L_{k} & =  \sum_{k=1}^{K} \mathop{\sum\sum}_{i,j\in C_{k}}(z_{j} - z_{i})^{2},\\
        & = J_{1} - \frac{1}{2} J_{2}.\\
        \end{split}
\end{equation*}
So we obtain \\
\begin{equation*} 
        \begin{split}
        L_{DCN} & =  L_{k} + \gamma L_{r},\\
        & = (1+\gamma) J_{1} - \frac{1}{2} J_{2} + \gamma J_{3}.\\
        \end{split}
\end{equation*}

\section{Proof of theorem 2}

The loss function $L_E(\phi(t))$ can be written as: \\
\begin{equation*} 
        \begin{split}
        L_E(\phi(t))  & \: = \: \sum_{i=1}^{N}\sum _{j=1}^{K} p_{ij}log(\frac{p_{ij}}{q_{ij}}) \ + \  \: \sum_{i=1}^{N}log(1-D_{\omega}(G_{\theta}(z_{i})))
        ,\\ & \: = \sum_{i=1}^{N}\sum _{j=1}^{K} p_{ij}log(p_{ij}) - p_{ij}log(q_{ij}) \ + \  \: \sum_{i=1}^{N}log(1-D_{\omega}(G_{\theta}(z_{i})))
        ,\\ \frac{\partial L_E(\phi(t))}{\partial z_{i}} & \: = \: \left [\sum _{j=1}^{K} \frac{\partial p_{ij}log(p_{ij})}{\partial z_{i}} - \frac{\partial p_{ij}log(q_{ij})}{\partial z_{i}}  \right ] + \  \ \frac{\partial log(1-D_{\omega}(G_{\theta}(z_{i})))}{\partial z_{i}} 
        ,\\ & \: = \left [\sum _{j=1}^{K}  -p_{ij} \frac{\partial log(q_{ij})}{\partial z_{i}}  \right ] + \  \ \frac{\partial log(1-D_{\omega}(G_{\theta}(z_{i})))}{\partial z_{i}}.
        \end{split}
\end{equation*}

Then, we compute $\frac{\partial log(1-D_{\omega}(G_{\theta}(z_{i})))}{\partial z_{i}}$ and $\frac{\partial log(q_{ij})}{\partial z_{i}}$ separately.
\begin{equation*} 
        \begin{split}
        &\frac{\partial log(1-D_{\omega}(G_{\theta}(z_{i})))}{\partial z_{i}}  \: = \: -\frac{(JG_{\theta}(z_{i}))^{T} \: .\:\triangledown D_{\omega}(G_{\theta}(z_{i}))}{1 - D_{\omega}(G_{\theta}(z_{i}))}.\\
        \frac{\partial log(q_{ij})}{\partial z_{i}} & \: = \: \frac{\partial log(\frac{(1 + \| z_{i} - \mu_{j}  \|^{2})^{-1}}{\sum_{{j}'} (1 + \| z_{i} - \mu_{{j}'} \|^{2})^{-1}})}{\partial z_{i}},\\
         & \: = \frac{\partial log((1 + \| z_{i} - \mu_{j}  \|^{2})^{-1})}{\partial z_{i}} - \frac{\partial log(\sum_{{j}'} (1 + \| z_{i} - \mu_{{j}'}  \|^{2})^{-1})}{\partial z_{i}},\\
        & \: = -\frac{2(z_{i} - \mu_{j})}{1+\left \| z_{i}-\mu_{j} \right) \|^{2}}+ \frac{2\sum_{{j}'} (z_{i} - \mu_{{j}'})(1 + \| z_{i} - \mu_{{j}'} \|^{2})^{-2}}{\sum_{{j}'} (1 + \| z_{i} - \mu_{{j}'} \|^{2})^{-1}}, \\
        & \: =-\frac{2(z_{i} - \mu_{j})(1+\left \| z_{i}-\mu_{j} \right) \|^{2})^{-2}}{q_{ij}\sum_{{j}'} (1 + \| z_{i} - \mu_{{j}'} \|^{2})^{-1}}
        + \frac{2\sum_{{j}'} (z_{i} - \mu_{{j}'})(1 + \| z_{i} - \mu_{{j}'} \|^{2})^{-2}}{\sum_{{j}'} (1 + \| z_{i} - \mu_{{j}'} \|^{2})^{-1}}.
        \end{split}
\end{equation*}

After substitutions, we have:

\begin{equation*} 
        \begin{split}
        \frac{\partial L_E(\phi(t))}{\partial z_{i}}  &\: =  \: -2\sum _{j=1}^{K} \frac{ \: p_{ij}\:(z_{i} - \mu_{j})(1+\left \| z_{i}-\mu_{j} \right) \|^{2})^{-2}}{q_{ij}\sum_{{j}'} (1 + \| z_{i} - \mu_{{j}'} \|^{2})^{-1}} + 2\sum _{j=1}^{K} \frac{ \: p_{ij}\: \sum_{{j}'} (z_{i} - \mu_{{j}'})(1 + \| z_{i} - \mu_{{j}'} \|^{2})^{-2}}{\sum_{{j}'} (1 + \| z_{i} - \mu_{{j}'} \|^{2})^{-1}} \\ 
        & \: +  \frac{(JG_{\theta}(z_{i}))^{T} \: .\:\triangledown D_{\omega}(G_{\theta}(z_{i}))}{1 - D_{\omega}(G_{\theta}(z_{i}))}, \\
        &\: =  -2\left [  \sum _{j=1}^{K} \frac{ \: p_{ij}\:(z_{i} - \mu_{j})(1+\left \| z_{i}-\mu_{j} \right) \|^{2})^{-2}}{q_{ij}\sum_{{j}'} (1 + \| z_{i} - \mu_{{j}'} \|^{2})^{-1}}\right ] + 2 \frac{ \: \sum_{{j}'} (z_{i} - \mu_{{j}'})(1 + \| z_{i} - \mu_{{j}'} \|^{2})^{-2}}{\sum_{{j}'} (1 + \| z_{i} - \mu_{{j}'} \|^{2})^{-1}}\\
        & \: +  \frac{(JG_{\theta}(z_{i}))^{T} \: .\:\triangledown D_{\omega}(G_{\theta}(z_{i}))}{1 - D_{\omega}(G_{\theta}(z_{i}))}, \\
        &\: =  -2\left [  \sum _{j=1}^{K} \frac{ \: p_{ij}\:(z_{i} - \mu_{j})(1+\left \| z_{i}-\mu_{j} \right) \|^{2})^{-2}}{q_{ij}\sum_{{j}'} (1 + \| z_{i} - \mu_{{j}'} \|^{2})^{-1}}\right ] + 2 \left [ \sum_{j=1}^{K}\frac{ q_{ij} (z_{i} - \mu_{j})(1 + \| z_{i} - \mu_{j} \|^{2})^{-2}}{q_{ij}\sum_{{j}'} (1 + \| z_{i} - \mu_{{j}'} \|^{2})^{-1}}\right ]\\ 
        & \: +  \frac{(JG_{\theta}(z_{i}))^{T} \: .\:\triangledown D_{\omega}(G_{\theta}(z_{i}))}{1 - D_{\omega}(G_{\theta}(z_{i}))},\\
        &\: =  -2\left [  \sum _{j=1}^{K}  \: p_{ij}\:(z_{i} - \mu_{j})(1+\left \| z_{i}-\mu_{j} \right) \|^{2})^{-1}\right ] + 2 \left [ \sum_{j=1}^{K}q_{ij} (z_{i} - \mu_{j})(1 + \| z_{i} - \mu_{j} \|^{2})^{-1}\right ] \\
        & \: +  \frac{(JG_{\theta}(z_{i}))^{T} \: .\:\triangledown D_{\omega}(G_{\theta}(z_{i}))}{1 - D_{\omega}(G_{\theta}(z_{i}))},\\
        &\: =   2 \sum _{j=1}^{K}  \: (1+\left \| z_{i}-\mu_{j} \right) \|^{2})^{-1}.(p_{ij}-q_{ij}).\:(z_{i} - \mu_{j}) +  \frac{(JG_{\theta}(z_{i}))^{T} \: .\:\triangledown D_{\omega}(G_{\theta}(z_{i}))}{1 - D_{\omega}(G_{\theta}(z_{i}))}.\\
        \end{split}
\end{equation*}

\section{Proof of theorem 3}

$\mu_{j}$ and $-z_{i}$ play symmetric roles in the first part of $L_E(\phi(t))$, and the regularization part does not depend on $\mu_{j}$. Therefore,
\begin{equation*} 
        \begin{split}
        \frac{\partial L_E(\phi(t))}{\partial \mu_{j}}  &\: = \: \frac{\partial L_E(\phi(t))}{\partial (-z_{i})} -  \frac{(JG_{\theta}(z_{i}))^{T} \: .\:\triangledown D_{\omega}(G_{\theta}(z_{i}))}{1 - D_{\omega}(G_{\theta}(z_{i}))}, \\
        &\: = \: - \frac{\partial L_E(\phi(t))}{\partial z_{i}} -  \frac{(JG_{\theta}(z_{i}))^{T} \: .\:\triangledown D_{\omega}(G_{\theta}(z_{i}))}{1 - D_{\omega}(G_{\theta}(z_{i}))},\\
        &\: = \: -2 \sum _{j=1}^{K}(1+\left \| z_{i}-\mu_{j} \right) \|^{2})^{-1}(p_{ij}-q_{ij})(z_{i}-\mu_{j}).
        \end{split}
\end{equation*}

\end{document}